\documentclass{report}

\usepackage[english]{babel} 
\usepackage[utf8]{inputenc}
\usepackage{amsfonts}
\usepackage{amsmath}
\usepackage{graphicx}
\usepackage{hyperref}
\usepackage[mathlines]{lineno}
\usepackage{mathtools}
\usepackage{cleveref}

\usepackage{tabularx}
\usepackage{subfig}
\usepackage{multirow}
\usepackage{array}
\usepackage{float}
\usepackage{tabularx}
\usepackage{stackengine}

\usepackage{natbib}
\DeclarePairedDelimiter{\ceil}{\lceil}{\rceil}

\title{%
Adversarial Regression \\
 \large Generative Adversarial Networks for Non-Linear Regression: \newline Theory and Assessment
 }

\author{Yoann Boget}
\date{August 2019}

\usepackage{graphicx}
\usepackage{appendix}
\usepackage{breqn}
\usepackage[ruled, vlined]{algorithm2e}
\SetKwInOut{Parameter}{parameter}

\begin{document}

\begin{titlepage}
   \begin{center}
       \vspace*{1cm}
        \LARGE
       \textbf{Adversarial Regression}
 \large       
       \vspace{0.5cm}
       
        Generative Adversarial Networks for Non-Linear Regression: \newline Theory and Assessment
 
       \vspace{1.5cm}

       \textbf{Yoann Boget}
       
        \vfill

        Supervisor at SLAC National Accelerator Laboratory: 
        
        Dr. Michael Kagan
        
        \vspace{0.4cm}
        Thesis Advisor at University of Neuchâtel:
        
        Dr. Clément Chevalier
        
 \vspace{1.5cm}

  \normalsize
       A thesis presented for the degree of\\
       Master in Statistics
 
       \vspace{0.4cm}

       Institute of Statistics\\
       University of Neuchâtel\\
       Switzerland\\
       August 2019
 
   \end{center}
\end{titlepage}

\tableofcontents

\maketitle

\chapter*{Prediction uncertainty}
\addcontentsline{toc}{chapter}{Introduction: Prediction uncertainty}

Prediction is one of the main issues in statistics. The problem can be exposed as follow. The variable $y$ is containing some information we want to predict from observations $X$. We assume that there is a true model, where $y$ can be expressed as a function $\phi$, of given data $X$, with some parameters $\theta$, and an random error term $\varepsilon$ and that can be written: \begin{equation}y= \phi\left(X, \theta, \varepsilon\right).\end{equation}.

In statistics, a difference is often made between forecast and prediction. Forecasting is most used in time series, where the observations are not conditionally independent. In the opposite, predicting imply the error terms to be independent. We will work with this background. By consequence, it will be assumed that the observations are conditionally independent so that the data can be factorized. In particular, we have:   \begin{equation} p\left( y \middle \vert \theta, X \right) = p\left(y_1, y_2, ... , y_n \middle \vert \theta, X \right)= \prod_{i=1}^n p\left(y_i\middle \vert \theta, X \right).\end{equation}

In the machine learning community, a distinction is made between regression and classification. Regression refers only to the cases where the dependent variable is continuous, while classification is used when the dependent variable is categorical. This work will focus on the regression case so that $y$ will be continuous. Without further information, $y$ will also be a one-dimensional variable, implying $y \in \mathbb{R}^n$, $n$ being the number of observations \footnote{In this chapter, we will use the statistical notation, following the notation in \citep{Tille}}.   

Within this background, prediction is an estimate containing information about $p_r(y\star \vert X\star)$, which is a probability density function. In many cases, the distribution is estimated through a single point estimate. Some very common ways to obtain a point estimate for $y\star$ is to compute the Maximum Likelihood (MLE), the Ordinary Least Squares (OLS) or the Maximum A Posterior (MAP) estimate of the parameters and estimate $y\star$ using the expectation of $y$ given those estimates. The principle behind the point estimate is to obtain a value representing our "best guess". Even without considering all the questions about how to model the "best guess", the point estimate gives no indication about how much guessing and how much confidence there is behind this "guess". The variance of the same estimate can be close to zero or to infinite, such that the same "best guess" can be with high confidence very close from the true value or almost a random guess. In the case where the distribution of the new observation is bi-modal, point estimates, whether the mean, the median or the mode does not bring useful information.

In many fields and especially when the aim is to evaluate risk, the "best guess" is not enough. We can think about climate change, health diagnostics, insurance, finance, etc. In all those fields, a "best guess" is not enough. At least, we need to know how likely the "best guess" is. In many application, the full probability density distribution for a new prediction is of major interest. We will refer to it as predictive uncertainty quantification. 

In the last few years, different methods able to produce predictive uncertainty quantification have been developed. In particular, some major breakthroughs have been done in the machine learning community. New types of methods have been implemented such as adversarial learning, Variational Auto-Encoders and and Bayesian Neural Networks. These methods can produce an approximation of the prediction distribution. In this work, we will focus on adversarial methods. In particular, we will show how to use Generative Adversarial Networks (GANs) \citep{goodfellow2014a} for regression and use it to build density estimates for the prediction distribution. We will call this adversarial regression. The aim will be then to explore the quality of the predictive uncertainty quantification.  

\chapter{Regressions}

The general model  $y= \phi\left(X, \theta, \varepsilon\right)$ with $y$'s being  independent and continuous is what we call the regression model. In this section, we will review the classical frameworks to deal with regression but, first, we remind the general issues of regression. 

We have $n$ realizations of an event: the data $\left(X, y\right)$. We assume they come from an unknown true joint distribution. We write: $(X, y) \sim p_r(X, y)$, where $p_r$ is the true joint probability distribution of $X$ and $y$. We want to use the information contained in one variable, say $X$ to produce estimation(s) about $y$. 

The issues come essentially from the fact that the true model $y= \phi\left(X, \theta, \varepsilon\right)$ is and remains unknown.   The aim is to estimate the conditional distribution  $p\left(y \vert X\right)$ or some properties of it. 

The main goal behind regression is usually of two types: inference and prediction. Here, we are particularly interested in prediction since, eventually, the aim is to approximate $p(y_{\star} \vert X_{\star}; X, y)$, where  $y_\star$ is a new prediction for a given a new observation $X_\star$\footnote{The expression "new observation" and "new prediction" means that the variables were not included in the original data. Consequently, the model is not built with these data. Also, the true value of $y$ for the prediction $y_{\star}$, if this expression has a meaning, remains unknown.}.

Let have a quick overlook at the main issues at which regressions are facing, starting from the model equation $y= \phi\left(X, \theta, \varepsilon\right)$. All the terms in the right-hand side arise difficulties, which can be distinguished in three main issues: unknowns, randomness, and complexity. 

First issue: almost everything is unknown:  $\phi$, $\theta$ and $\varepsilon$ are unknown. This lead to an infinity of solutions. As we are going to see, the solution about $\phi$ will be to restrict it to a family of functions with a fixed number of parameters. 

Second issue: the model is random. The error term represents the intrinsic stochasticity in the model. Fortunately, we can assume the expectation of the error term to be zero ($E(\varepsilon)=0$. If it was not the case, we could find another function and/or other parameters to fit the data better and have $E(\varepsilon)=0$. 
However, the stochasticity of the model is terrible. It means that the data, the only thing we have, are random. By consequence, the estimates of the parameters are going to be random too. It also means that there is an
irredeemable uncertainty of the prediction. Moreover, the error term is random and unknown, so that we can only estimate it. 

Third issue: the complexity of the true model. Complexity can come from $\phi$ that can be any kind of functions, including function without a closed-form. For the same reason the dimension $p$ of $\theta$, which is a vector in $\mathbb{R}^p$, can be arbitrarily large. In the same vein, the matrix of independent variables $X \in \mathbb{R}^{n \times m}$ can also be arbitrarily large in both dimensions, $n$ the number of observations and $m$ the number of variables. If a large number of observations is usually not an issue but computational, a large number of variables prevents the design matrix to be full rank which makes $X^T X$ not invertible. Depending on the choice of the model, the non-invertibility of $X^T X$ can be damageble.

In this chapter, we will succinctly present three main approaches to address regression and expose how they have to deal with the above issues.  Starting from the classic formulation of linear regression, we are going to see what brings the Bayesian framework and neural networks. Simultaneously, this presentation will be an opportunity to introduce the theoretical background and basic techniques, we are going to use. 

\section{Classical frameworks for Regressions}

First, we review the most classic form of prediction in statistics: classical least square linear regression. The idea is to express $y$ as a linear combination of $X$ and an error term, which is assumed to be normally distributed. Thus, the general linear regression model can be written as follow: 

\begin{equation} y = X\beta + \varepsilon, \end{equation}

where $y$ is a vector in $\mathbb{R}^n$,

X is a full rank matrix in $\mathbb{R}^{n \times p}$, 

$\beta$ is a vector in $\mathbb{R}^p$ and 

$\varepsilon \  \sim \mathcal{N}(0,\,\sigma_{\varepsilon}^{2}I_n)$, $I_n$ being and $n \times n$ identity matrix. 

This can be read like a list of restriction and additional assumption in comparison to the general model $y= \phi\left(X, \theta, \varepsilon\right)$. $\phi$ is restricted to a linear combination between $X$ and $theta$ and the error term has no relation with any other terms and is normally distributed. Finally, $X$ has to be full rank.

Since, $\phi$ is given, the remaining task consists of estimating $\beta$ and $\sigma_{\varepsilon}^{2}$. In this framework, the estimation can be done by minimizing the least square error. The solution has a closed form given by the formula: \begin{equation}\hat{\beta} = (X^TX)^{-1}X^Ty\end{equation}
and \begin{equation} \hat{\sigma_{\varepsilon}^{2}}=\frac{e^Te}{(n-p)}\end{equation}
where, $e= y-X\hat{\beta} \in \mathbb{R}^n$ is the vector of residuals.  
It is obvious that $\hat{\beta}$  and $\hat{\sigma_{\varepsilon}^{2}}$ are random since they depend on $y$. With those two estimates, it is possible to compute a prediction for a set of new explanatory variables. We have: \begin{equation}\hat{y_\star}= X_{\star} \hat{\beta}.\end{equation}
The expected error is equal to zero: $E(\hat{y_{\star}}-y_{\star})=0$ and the error variance can be estimated as follow: $\hat{var}(\hat{y_{\star}}-y_{\star})= \hat{\sigma_{\varepsilon}^{2}}(x_\star(X^TX)^{-1}x_\star^T + 1)$. It is thus possible to build a confidence interval for the predictions. 

All the demonstrations appear in the Lecture Notes of \textit{Advanced Regression Method} given by Pr. Yves Tillé \citep{Tille}. Since it is not central for the demonstration, we do not reproduce them here. 

As such, linear regression offers prediction and confidence interval for them. The model has one major benefit: its simplicity. Unfortunately, this simplicity has a cost:
\begin{enumerate}
\item The model requires to assume the normality of the error term, which is a strong assumption that does not hold in many cases. 
\item The model has to be a linear combination of the input variables. Again, it is a very strong restriction over the possible function, even if this issue can be partially avoided by working with transformed variables (i.e. with $\tilde{X}= \Phi(X)$ where $\Phi$ is an arbitrarily complex function). 
\item High dimensional inputs can cause problems concerning the required condition for the design matrix to be full rank.
\item It is possible to build a confidence interval for the new predictions. However, it does not provide an estimate of the probability density for the new prediction.
\end{enumerate}

Of course, the so-called Generalized Linear Model (GLM) proposes some (partial) solutions, mainly concerning the first of the above issues. While we will not discuss GLM in detail here, we will present the basic concepts of logistic regression, since they are strongly connected with some techniques we are going to use. 

\subsection{Logistic Regressions}

Logistic regressions are famous to deal with categorical response variables and it is also in that purpose we are interested in. Classifier neural networks can be interpreted as non-linear logistic regressions. Indeed, the last layer of such classifiers is logistic regression. Even if we are mainly interested in regression rather than in classification, some generative adversarial networks include a classifier. Hence, it is of interest to review logistic regression to understand this kind of networks. 

\subsubsection{Binary logistic regression}

First, we have a look at the binary logistic regression. The response variable can only take two value: 0 or 1. 
Let define $Z$ as the result of a linear combination of the matrix X so that $Z = \theta^T X$. 
Then, the sigmoid function $\sigma\left(Z\right)$ is a function from $\mathbb{R}$ to $\left[0, 1\right]$ defined as follow: 
\begin{equation}\sigma\left(Z\right)= \frac{exp\left(Z\right)}{1+exp\left(Z\right)} = \frac{1}{1+exp\left(-Z\right)} = \sigma\left(Z\right) = \hat{y} \end{equation}
where $\hat{y}$ is interpreted as the prediction probability for $Z$ being of class 1.
It follows that we have:
\begin{equation} Z= \theta^T X = log\left(\frac{\sigma\left(Z\right)}{1-\sigma\left(Z\right)}\right).\end{equation}
Since, the quantity $\hat{y}=\sigma(Z)$ gives the probability of being of class 1 and the model follows a Bernoulli distribution, the likelihood is given by: 
\begin{equation}p\left(y \vert \theta\right) = \prod_{i=1}^n \left(\sigma\left(Z\right)\right)^{y_i}\left(1-\sigma\left(Z\right)\right)^{1-y_i}.\end{equation}
The usual error function for the likelihood is the negative log-likelihood known as the cross-entropy loss function:
\begin{equation} L\left(y, \theta\right) =  -log\left( p\left(y \vert \theta\right)\right)= \sum_{i=1}^n y_i \cdot log\left(\sigma\left(Z\right)\right) + \left(1-Y_i\right)log\left(1-\sigma\left(Z\right)\right).\end{equation}

This result is very important for the neural network. It is used in many binary classification tasks. 
\subsection{Multinomial logistic regression}

The other main classification task is quite similar but with more than 2 classes. Let denotes $J$ the number of output classes. $Z$ is now a vector of size $J$ ($Z \in \mathbb{R}^J$), $Z_i$ denoting its $i^{th}$ component.   
\begin{equation} \sigma(Z_i) = \frac{e^{Z_i}}{\sum_{j =1}^{J}e^{Z_j}}. \end{equation}
Following the same principle as the binary logisitic regression, $\sigma(Z_i)$ represent the probability for $Z_i$ of being of class $i$. It is easy to see that the binary case is just a particular case of a multinomial regression when $J=2$. The general cross entropy loss function take this form:

\begin{equation}L(y_i) = -\sum_{j=1}^J \mathbf{1}_{\{y_i=j\}}\log(\hat{y_i}).\end{equation}

\section{The Bayesian approach}

We introduce briefly the Bayesian framework. The Bayesian version of the linear regression offers the possibility to compute a probability density prediction for a new observation, which is an example of what we are looking for. 
 
The Bayesian approach\footnote{This section is based on references about Bayesian statistics \cite{Albert, Chevalier} and Bayesian methods applied to machine learning \cite{Murfy, barberBRML2012, Bishop}.}  aims to model knowledge about some events in terms of probability. Consequently, the model parameters are no longer fixed unknown values but random variables. This change of perspective makes possible to compute the probability density for the parameters given the data and, consequently, the probability density for a new prediction. Typically, we are interested in the probability density function of the parameter given the data: $ p\left( \theta \middle \vert Y\right)$. Thanks to Bayes Theorem, we have the following relation: \begin{equation}  p\left( \theta \middle \vert Y\right) =  \frac{p\left( Y \middle \vert \theta\right) p(\theta)}{p(Y)}. \end{equation}

On the right-hand side of the equation, we have three terms. The first one $p\left(Y  \middle \vert \theta \right)$ is the likelihood. The second term in the equation, $p(\theta)$, is the prior and represents the probability distribution over the parameters from prior knowledge before having the data. The third one $p(Y)$ is a constant in order that the probabilities integrate to one \begin{equation} p(Y)= \int p\left( Y \middle \vert \theta \right) p(\theta)d\theta .\end{equation} However, an expression proportional to the posterior is often sufficient, so that only the numerator is computed. In that case, we write: \begin{equation}p\left( \theta \middle \vert Y\right) \propto p\left(Y \middle \vert  \theta \right) p(\theta).\end{equation}

 The Bayesian approach offers a theoretical answer to the uncertainty issue about prediction. The full distribution for a new prediction is definitively a satisfying answer to uncertainty. Unfortunately, the posterior distribution over the parameters is practically rarely tractable. In high dimensional non-linear spaces, it is almost never the case.

\subsection{Bayesian linear regression}
Since being Bayesian offers a convincing framework to deal with uncertainty, it is of interest to go through the same simple linear regression model and see how the Bayesian approach deals with it.   

The likelihood remains the same. It is normally distributed with mean $X\beta$ and variance $ \sigma^{2}$. Since we are interested to express 
\begin{equation}p(y \vert X, \beta, \sigma^{2}) = \mathcal{N}(y \vert  X\beta  ,\sigma I_N) = ((2\pi) \sigma^2)^{-\frac{n}{2}} exp(-\frac{1}{2\sigma^2}(y-X\beta)^T(y-X\beta)) ,\end{equation} $I_N$ is an identity matrix of size $n$. We want to compute the posterior distribution over the parameters $\beta$ and $\sigma^2$. This requires to define a prior on the parameters $\theta$ and $\sigma_{\varepsilon}^{2}$. Of course, all kinds of priors are allowed and possible. However, most of them are intractable and require numerical approximation. Hence, it is very common to choose a so-called conjugate prior which allows an analytical result.
Actually, instead of choosing a prior for the joint distribution $p(\theta, \sigma^2)$, it is very common to separate the joint prior in two distributions, since we have: \begin{equation}p(\beta, \sigma^2) = p(\beta \vert \sigma^2)p(\sigma^2).\end{equation}
   
The solution for a conjugate prior is often called NIG (standing for Normal Inverse-Gamma) since the prior for $\beta$ would be a Normal distribution and the one for $\sigma^2$ an Inverse-Gamma. The posterior is obtained as follow:
\begin{equation}p(\beta, \sigma \vert X, y) = \mathcal{N}(y \vert X, \beta, \sigma^2) *  \mathcal{N}(\beta \vert \sigma^2; \beta_0, \Sigma_0) IG(\sigma^2; a_0, b_0)\end{equation}
where $\beta_0$, $\Sigma_0$, $a_0$, $b_0$ are hyperparameters for the prior distributions.  
Eventually, the posterior can be express as a NIG: 
\begin{equation}p(\beta, \sigma \vert X, y) = NIG(\beta, \sigma^2 \vert \beta_n, \Sigma_n, b_n, a_n)\end{equation}

where $\beta_n = \Sigma_n(\Sigma_0^{-1}\beta_0 + X^Ty)$,

$\Sigma_N = (\Sigma_0^{-1} + X^TX)^{-1}$, 

$a_n = a_0 + \frac{n}{2}$,

$b_n = b_0 + \frac{1}{2}(\beta_0^T \Sigma_0^{-1} \beta_0 + y^T y - \beta_n^T \Sigma_N^{-1} \beta_n) $. 

That is the very general form. Some of the prior's parameters are particularly popular. For instance, a non-informative prior can be obtained by setting $\beta_0=0$, $\Sigma_0= \infty I$, $a_0=\frac{-p}{2}$, $b_0=0$, which gives $p(\beta, \sigma^2)= \sigma^{-(p+2)}$, $p$ being the number of columns of $X$. An other example arise keeping the uninformative prior for $\sigma^2$ and choosing a prior with the form  $\mathcal{N}(0 , \tau^2 I)$ for $\beta$. It gives a Maximum A Posteriori (MAP) equivalent to the estimate obtained with classic ridge regression, having $\lambda = \frac{\sigma^2}{\tau^2}$. 

Moreover, it can be shown that the MAP for $\beta$ with an uninformative prior turns out to yield equivalent result as the frequentist approach. Not only the parameters are the same. The frequentist $1-\alpha$ confidence interval is also the same as the $1-\alpha$ credible interval from the Bayesian perspective. 

The interest of being Bayesian is not obvious in this case. It provides similar results as the classical linear regression. The restriction about the model family and the required additional assumptions are the same as the one made in the frequentist framework. Nonetheless, the method changes the perspective. Instead of having a point estimate and a confidence interval, we now have a probability distribution for the new prediction, which is what we are looking for.

\section{Neural networks}

Neural networks for regression are important for two reasons. First, they release considerably the restrictions and the assumptions of the previous methods. Second, the generative adversarial networks (GAN) are neural network based. 

Neural networks for regression tasks can be seen as a proposition to address the restrictions and additional assumptions of the linear models. First of all, neural networks enlarge drastically the family of functions take into consideration. They are chains of nested functions including non-linear transformation able to approximate almost any kind of real-valued function. The only restriction they have about the size of the design matrix is the computational power at disposal. They make no assumption about the error term.
However, basic regression neural networks have two important limitations. First, they often overfit the data, implying regularization techniques and hyperparameters to prevent it. Second, they offer only point estimate and provide no information about the prediction uncertainty. 

In addition, Neural Networks are also important for this study. The adversarial regression that we are going to test is based on neural networks. So, we can introduce Neural Networks in the same time for their ability to perform regression and as a method, we are going to use.  

In their simplest form, neural networks can be thought of as a non-linear multi-regression model. Indeed, a neural network consists of an input layer containing the data, one or several hidden layers and an output layer, each layer containing several neurons also sometimes called nodes. The number of units in each layer and the total numbers of layers can be seen as hyperparameters. We will come back later to this. 

\subsection{Feedforward the neural network}

Each unit is a place where two operations happen. First, each unit of each layer computes a linear combination of the outputs of the previous layer. The result for the $j^{th}$ unit in layer $l$ is noted $z^{[l]}_j$. Therefore:

\begin{equation} z^{[l]}_j=(w^{[l]}_{j,.})^T a^{[l-1]}+b^{[l-1]}_j.\end{equation}

Second, the linear combination $z^{(i)[l]}$ is then transformed by a non-linear activation function $ g^{[l]} $. Consequently, the output $a^{[l]}_j$ of the $j^{th}$ unit is obtain as follow: \begin{equation}a^{[l]}_j= g(z^{[l]}_j)= g[ (w^{[l]}_{j,.})^T a^{[l-1]}+b^{[l-1]}_j ] \end{equation}

\begin{figure}[ht!]
\centering
\includegraphics[scale=0.5]{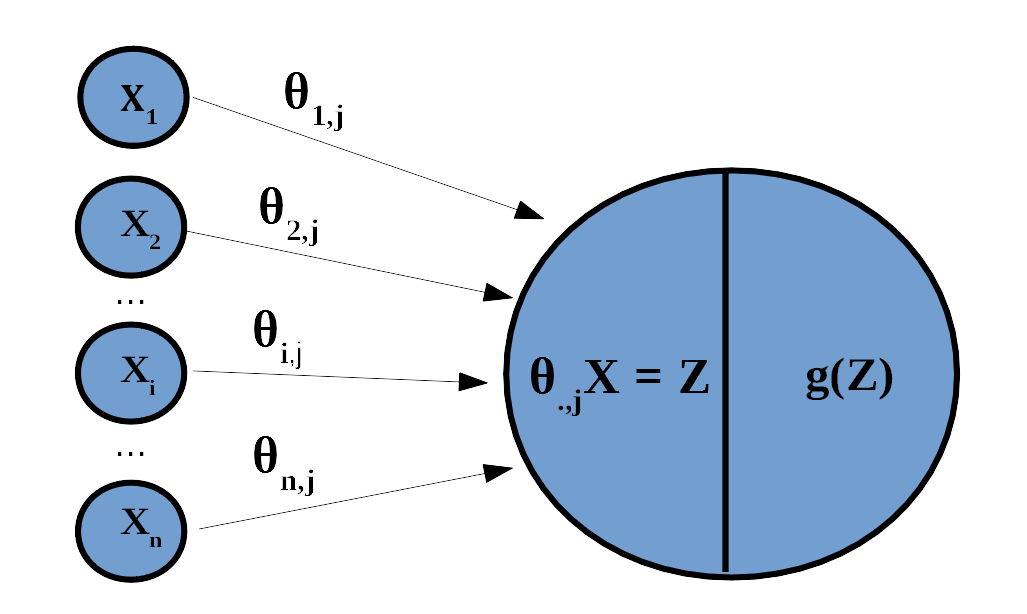}
\caption{Figure 1}
\label{fig:figure 1}
\end{figure}

Figure \ref{fig:figure 1} is a graphical representation of a single unit.

Once, we have understood how does work a single unit, the understanding of the whole network is straight forward since the neural network is simply a succession of layers, each one containing a fixed number of units. The computation of layer $l$ is given as follow: 

\begin{equation} A^{[l]}= g(Z^{[l]})= g[ (W^{[l]})^T A^{[l-1]}+b^{[l-1]} ] .\end{equation}

We will come back later on the issue of the initialization of the weights but let assume for now we have set some weights. The forward pass is very easy. The pseudo code holds in 4 lines: 

So, the model is only a succession of linear combinations followed by a non-linear activation function.

\subsection{ The activation functions}
Why do we need an activation function? It can be proven that a linear combination of a linear combination is still a linear combination. Therefore, without a non-linear activation function, a neural network would be equivalent to a linear combination of the input. Therefore, the activation function is essential. 
There are plenty of activation functions. We present here the most popular ones. 

The \textit{sigmoid} function has already have been seen in the binary logistic regression. 
In the machine learning context, the logistic function for multinomial logistic regression is called the \textit{softmax} function.
The rectified linear unit (Relu) function is probably the most famous and the most used in the hidden layers. Its expression is very simple:
\begin{equation}g(Z) = max(0, Z) .\end{equation}
The drawback of the Relu function is that its gradient is zero for all negative Z. To avoid it, a slightly modified version of it has been developed and called leaky Relu. Its expression is \begin{equation}g(Z) = max(aZ, Z) ,\end{equation} $a \in [0,1]$ and is usually small (typically $10^{-1}$ or $10^{-2}$). 

Thanks to the forward pass, we have now the matrix $\hat{Y}$.

\subsection{The loss functions}

The neural networks can be interpreted as a non-linear regression. As we would do in linear regression, the aim is to find the parameters or "weights" minimizing the cost function. Because of the complexity of the function to minimize, there is no closed form to find those minima. Nonetheless, the strategy will remain the same: compute an error and minimize it.
Therefore, we need a mathematical expression for the cost. Loss and cost functions refer to all the function able to compute the error. I present here two among the most famous loss function. 

The mean square error is the sum of the square of the differences between the predicted value and the true value divided by the number of observations.

\begin{equation} \mathcal{L}^{(i)}(\hat{y}^{(i)}, y^{(i)})= \sum_{i=1}^n (y_i -\hat{y_i})^2 
.\end{equation}

For classifier, the most common function is the cross-entropy loss:

\begin{equation} \mathcal{L}^{(i)}(\hat{y}^{(i)}, y^{(i)})= -\sum_{j=1}^J \mathbf{1}_{\{y_i=j\}}\log(\hat{y_i}),\end{equation} $J$ is the number of classes.

Of course, the loss function can take different expression depending on the aim. We are going to see different loss function in the chapter about GANs. 

\subsection{Backpropagation and gradient descent}
Neural networks aim to minimize the cost function. There are several optimization methods but all of them are improvements of the first general idea: gradient descent. Let present first gradient descent and then some of the most popular improvement of it.  
The general idea of gradient descent consists of gradually updating the parameters to find a minimum, i.e. some values for the parameters where all partial derivatives with respect to the parameters are equal to zero. 
Once we have the values for all $Z^{[l]}$ and $A^{[l]}$ and we have the cost, we can use it to optimize the weights. 
The idea is to use gradient descent to reach a minimum.
Gradient descent works like this: 
\begin{itemize}
    \item Take the current values of the parameters as a starting point.
    \item Compute the gradient of the cost function for these values i.e. its first partial derivative with respect to the parameters. 
    \item Compute the gradient of the cost function for these values i.e. its first partial derivative with respect to the parameters.
    \item Repeat this process until the gradient converges around zero.
\end{itemize}{}

What a "small step" means remains unclear for now. Let call it the learning rate. We will come back later on this. 
So, the general formula for updating the parameter is as follow: 
\begin{equation} \theta_i = \theta_{i-1} - \alpha \frac{\partial J_{\Theta_{i-1}} }{ \partial \theta_{i-1} }\end{equation}
where $\alpha$ is the learning rate, $J_{\Theta_{i-1}}$ is the cost computed with the former parameters, $\theta_{i-1}$ represents the parameter before the update and $\theta_i$ the same parameter updated.
If the learning rate $\alpha$ is small enough, then the cost will always decrease until it reaches a value close to a local minimum.

\subsection{The chain rule}
The gradient descent implies to compute the partial derivative of the cost function with respect to the parameters. It may look very difficult since we have non-linear functions with respect to many parameters but, thanks to the chain rule, it is easier than it may seem at first sight. 

The chain rule of derivation is a way to compute the derivative of variables inside nested functions. Let be $f$, $g$, $h$ some functions and $x$ a variable. The function $f$ is the outer function and $h$ is the inner function as follow: $f(g(h(x)))$. Then, the derivative of $f$ with respect to $x$ is the product of the derivative of $h$ with respect to $x$, of the derivative of $g$ with respect to $h$ and of $h$ with respect to f. We can write:
\begin{equation} \frac{df}{dx}= \frac{df}{dg} * \frac{dg}{dh} * \frac{dh}{dx}.\end{equation}

Since a neural network can be seen as a range of functions nested within each other, the chain rule does apply. 
So if we want to compute the derivative of the cost function with respect to the weight of last layer $W^{[L]}$, we have:
\begin{equation} \frac{\partial J }{ \partial W^{[L]}} = \frac{\partial J }{ \partial A^{[L]}}* \frac{\partial A^{[L]} }{ \partial  W^{[L]}  }. \end{equation}

We can apply the chain rule even further to compute the derivative of the previous layers. Following the same method, we can compute all the weights back to the first layer. A general formula of the previous example could be written like this: 
\begin{multline}
\frac{\partial J_{\Theta} }{ \partial W^{[l]}} = \frac{\partial J }{ \partial A^{[L]}}  * \frac{\partial A^{[L]} }{ \partial  Z^{[L]}  }*\frac{\partial Z^{[L]} }{ \partial  W^{[L]}  }*\frac{\partial W^{[L]} }{ \partial  A^{[L-1]}  }  * \frac{\partial A^{[L-1]} }{ \partial  Z^{[L-1]}} *\frac{\partial Z^{[L-1]} }{ \partial  W^{[L-1]}  }*\\
 ... * \frac{\partial Z^{[l+1]} }{ \partial  A^{[l]}} * \frac{\partial A^{[l]} }{ \partial  W^{[l]}} 
\end{multline}
The chain rule allows a relatively easy computation of the gradient for all the parameters so that eventually  they can be updated. Gradient descent do not ensure to reach the global minimum but a local minimum. In practice, however, it usually produces good results anyways. 

\subsection{Improvement of Gradient Descent}

Gradient descent is the principle of optimization. However, some improvements make the process faster. In this section, we will see some important improvements that we are going to use: stochastic and mini-batch gradient descent,  momentum, RMSprop, Adam. 

\subsubsection{Stochastic and Mini-batch Gradient Descent}
Stochastic gradient descent consists of drawing a random subset of the data and updating the parameters after each observation instead of the gradient on the whole dataset. The idea is that it is much faster to compute the gradient for one observation than for all the dataset. Even if a single update could yield a loose approximation of the loss, it is compensated by the fact that it is possible to compute much more iterations.

An alternative to a strict stochastic gradient descent is mini-batch gradient descent. Instead of taking only one observation the idea is to take a small batch of data. Again the data are first randomly sorted in mini-batches. The parameters are updated on each mini-batches.

We will call an epoch one full training cycle through the whole dataset so that an epoch of stochastic gradient descent consists of $n$ iterations and an epoch of mini-batches of size $m$ consists of $\ceil{\frac{n}{m}}$ iterations.

\subsubsection{Momentum}

Gradient descent with momentum is not a recent method. It was proposed in 1964 by \citet{Polyak1964}. It has been popularized again in the machine learning community by \citep{Sutskever2013}. The idea with momentum is to update the parameters not only with respect to the last computation of the gradient but taking into account an exponentially weighted (moving) average of the last iterations.

At iteration $i$, we compute $V_{d\theta}^{(i)}$ as a linear combination of the previous $V_{d\theta}^{(i-1)}$ and the new gradient $V(d\theta)$ so that 

\begin{equation}V_{d\theta}^{(i)} = \beta V_{d\theta}^{(i-1)}+ (1-\beta) d\theta  \end{equation}

where $\beta$ is a parameter controlling the weight of the past iteration relatively to the weight of the new gradient. A $\beta$ close to one gives much weight to the past, a $\beta$ close to zero gives bigger weight to the new gradient.
The parameters are not updated with $\partial\theta$ but with $V_{\partial\theta}^{(i)}$ as follow: 

\begin{equation} \theta_i = \theta_{i-1} - \alpha V_{d\theta}^{(i)}. \end{equation}

That way, the update of the gradient tends to continue in the direction given by the exponentially weighted average of the last iterations.

Momentum is particularly useful with stochastic or mini-batch gradient descent. Since one observation, or even an observation, yield to fluctuating gradient values because of the randomness of the process, the momentum compensates these fluctuations.  

\subsubsection{RMSprop}
The RMSprop is another technique for speeding the optimization. It has an unconventional background since it was first proposed in an online course and spread from there. The class is not available anymore but the slide notes are \citep{HintonG}. Instead of computing the moving average of the gradient, we compute the moving average of the second derivative. The average of the second derivative would tend to zero in the dimensions corresponding to the direction where the values of the parameters are moving during the last iterations and tend to infinity in the dimensions orthogonal to them. Just like with the momentum but with the second derivative, we compute an exponentially weighted average: 

 \begin{equation}S_{d\theta}^{(i)} = \beta S_{d\theta}^{(i-1)}+ (1-\beta) d\theta^2 \end{equation}.
 
 The parameters are then updated with the derivative weighted by the inverse of the square root of $S_{d\theta}^{(i)}$. Formally, the expression of the update si written: 
 
\begin{equation} \theta_i = \theta_{i-1} - \alpha \frac{d\theta}{\sqrt{S_{d\theta}^{(i)}}}. \end{equation}

That way the update is bigger in the dimensions parallels to the direction where the last update have been made. 

\subsubsection{Adam}

Adam optimizer, proposed in \citep{Kingma2014}, is actually the combination of gradient descent with momentum and RMSprop. $V_{d\theta}^{(i)}$ and $S_{d\theta}^{(i)}$ are computed the same way and the parameters $\beta$ are recalled respectively $\beta_1$, $\beta_2$. Also Adam use a corrected version of $S$ and $V$, reducing their weight in the beginning of the training. The corrected version is modified as follow: \begin{equation}V_{d\theta}^{Corr, (i)}= \frac{V_{d\theta}^{(i)}}{1-\beta^t}\end{equation} \begin{equation} S_{d\theta}^{Corr, (i)}= \frac{S_{d\theta}^{(i)}}{1-\beta^t},\end{equation} $t$ is the number of iterations.
Eventually, the parameters are updated as follow: 

\begin{equation} \theta_i = \theta_{i-1} - \alpha \frac{V_{d\theta}^{Corr, (i)}}{\sqrt{S_{d\theta}^{Corr, (i)}}}. \end{equation}

Adam has been considered as state-of-the-art. It is used in a lot of neural networks. However, as we are going to see, the algorithm is sometimes used with $\beta_1=0$ or $\beta_2=0$. In those cases, Adam is equivalent respectively to RMSprop or gradient descent with momentum.

\subsection{A powerful Algorithm}
Putting all together, the general algorithm for a standard basic neural network can be written like this:

\begin{algorithm}[H]
 \KwData{$X \in \mathbb{R^{m \times p} }$}
 \KwResult{$\hat{Y}$, Updated: W, b.}
 
 Initialize: W and b\; 
 Set: L, g()'s and $\alpha$ \; 
$A^{[0]}=X^T$ \;
 \While{ $\frac{\partial J_{\Theta} }{ \partial W} \not\approx 0$}  {
 \For{$l$ in 1:$L$}{
 $Z^{[l]}= W^{[l]T} A^{[l-1]}+b^{[l]}$\;
   $A^{[l]}=g( Z^{[l]} )$}
 $\hat{Y}=A^{[L]}$\;
 Compute: $J_{\Theta}(\hat{Y}, Y)$\;
 $dA{[L]} = \frac{\partial J }{ \partial A^{[L]}} $ \;
 \For{$l$ in $L$:1}{
  $dZ^{[l]}= dA^{[l]} g'(Z^{[l]})$\; 
 $dW^{[l]}=dZ^{[l]} A^{[l]T}$\;
$ db^{[l]}=  \sum_{i=1}^m dZ^{(i)[l]} $\;
 $dA^{[l-1]}= W^{[l]T}dZ{[l]} $\;
 }
 $W=W-\alpha \cdot dW$\;
 $b= b- \alpha \cdot db$
 }
 
 \caption{forward pass}
\end{algorithm}

Neural networks are very good at building complex models. Even very simple architecture can lead to astonishing results. With just a few layers, a neural network can recognize (predict) handwritten digits from a picture with very good accuracy. 
However, they have some important drawbacks too. First, they are computationally expensive. Neural networks have many parameters to train, i.e. to estimate through gradient descent. The gradient descent itself is an iterative process computationally expensive in its principle. Second, some neural networks tend to overfit the data. Fortunately, they exist several regularization methods. However, all those methods involve additional hyperparameter(s) that requires to be tuned. Third, neural networks demand a lot of data. It is another consequence of the high number of parameters. Efficient training of neural networks implies many training examples. Finally, basic neural networks do only provide point estimates. Since the prediction depends on parameters that suffer non-linear transformation and combination with other parameters it is impossible to retrieve the uncertainty neither from the data nor from the parameters.

\section{chapter conclusion}
The ideal regression methods would combine the best of the two: the power of neural networks and the ability to quantify uncertainty for new predictions. Generative adversarial networks are potentially able to put these qualities together. The aim would be to test how well they can do it. 

\chapter{From Generative Adversarial Networks to Adversarial Regression}
GANs are a class of algorithms aiming to "learn" any kind of distributions from a sample. "Learning" here means that these algorithms are able to generate new data which can be interpreted as coming from an approximation of the true model. Therefore, generating new data is similar to sampling from an approximation of the true distribution. The generated sample can be used to compute any kind of statistics. 

GANs are said to be adversarial because they can be presented as two neural networks playing against each other. The Generator generates fake data. The Discriminator learns to distinguish between real data and fake data. It is a simple classifier. It is a function from $\mathcal{X}$, the space spanned by the variables, to $[0, 1]$. The Generator tries to produce data that fool the Discriminator. It is a function from the latent space $\mathcal{Z}$ usually in $R^d$, where $d$ is an arbitrary but preferably large enough integer, to $\mathcal{X}$. We will discuss the details later. By iteration, the Generator produces eventually fake data indistinguishable from the real data. 

The general idea and the first GAN were presented in 2014 by \citeauthor{goodfellow2014a} in a paper simply entitled \textit{Generative Adversarial Network}. Later, Goodfellow wrote a pedagogical tutorial about GANs \citet{Goodfellow2016}. This first idea has been followed by numerous propositions to improve this idea. 

GANs are not models specifically made for regression. They are primarily generative models, i.e. models able to approximate any distribution and therefore generate new data from it. We will present later a conditional form of GAN to use it for regression, but first, let present the general formulation. 

\section{Theoretical background of GANs}

\begin{figure}[ht!]
\centering
\includegraphics[scale=0.5]{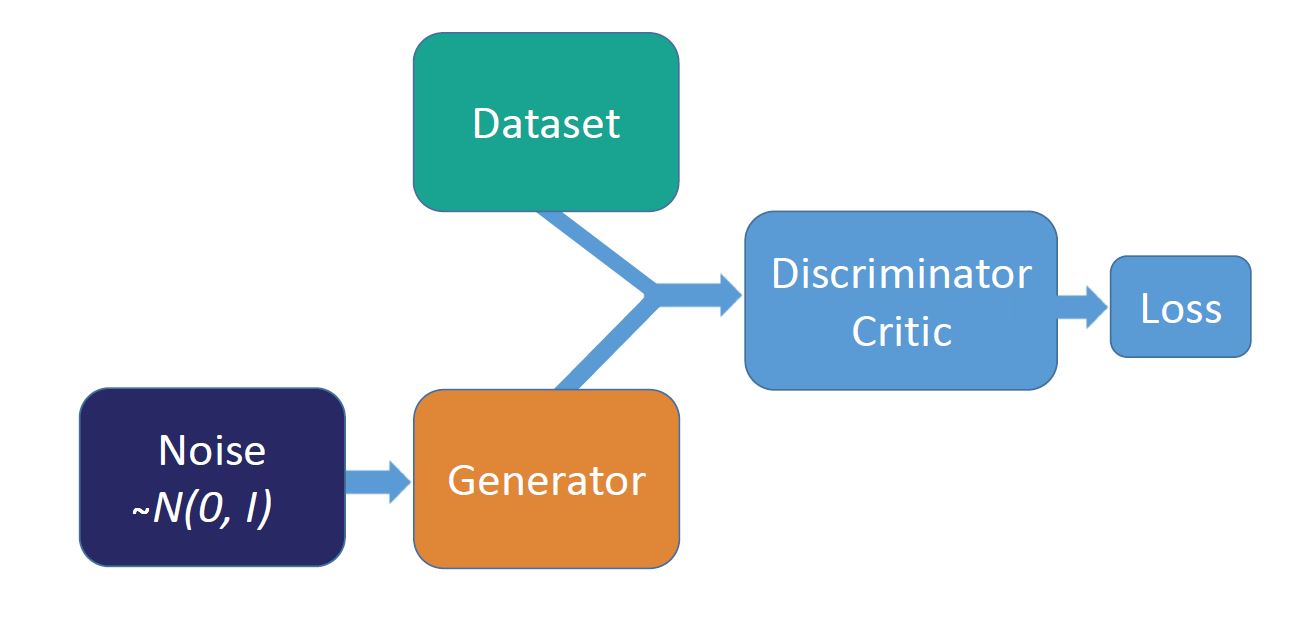}
\caption{Figure 1}
\label{GAN}
\end{figure}

We present in the section the main ideas of \citep{goodfellow2014a}. We have an unknown density function $p_{r}(x)$ and a dataset containing realizations of it. We want to use this sample to be able to produce new realizations of an approximation of $p_{r}(x)$.
The idea of GAN is as follow:
\begin{enumerate}
\item We generate some data $Z$ from a simple known distribution, typically the standard normal.
\item We train a Generator to simulate as good as possible realizations of $X$, so that we have $G(z)=x \sim p_g(x)$.
\item We train a Discriminator $D$ to distinguish between $x$ coming from $p_r(x)$ and the one coming from $p_g(x)$
\end{enumerate}
Actually, $G$ and $D$ are trained simultaneously through optimization algorithm each of them making one step of optimization after the other. The process is as a game between two players \footnote{Note: In fact, it is related to the game theory and Nash equilibrium. For details, see \citep{Oliehoek2018}}. The first player the Generator try to fool the Discriminator by generating sample as close as possible of the real distribution. The Discriminator tries not to be fooled. The better one player becomes, the better the other one has to be to succeed in its task. Figure \ref{GAN} illustrates the architecture of a GAN (depending on the loss function the Discriminator is recalled Critic). 

\subsection{The Discriminator} 

Let start with the discriminating part. It is a classification task. As shown previously, the loss function for such a case is given by the cross entropy loss function. Plugging in the GAN's notation, the loss function can be written: 
\begin{equation} \max\limits_D V(D, G) = \mathbb{E}_{x\sim p_{r}(x)} [log(D(x))] + \mathbb{E}_{z\sim p_{z}(z)} [log (1-D(G(z)))] .\end{equation}

Here, we can apply a change of variable that is not obvious if we do not want to assume that G is invertible (which is not generally the case with neural networks). As emphasized by \citep{RomeScott} and thanks to the Radon–Nikodym theorem, it is possible to write: 
\begin{equation} \max\limits_D V(D, G) = \mathbb{E}_{x\sim p_{r}(x)} [log(D(x))] + \mathbb{E}_{x\sim p_{G}(x)} [log (1-D(x))] .\end{equation}
For a fix G(z), we can then find the $D^{\star}_G$ that maximize the loss function: 

\begin{equation} D_G^{\star}= \text{arg}\max\limits_D \int p_{r}(x) [log(D(x))] + p_{g}(x) [log (1-D(x))] dx.\end{equation}

Since it is the same as maximizing the integrand, the maximum is obtained where the first derivative is null if the second derivative is negative for all $x$'s. 
\begin{equation}
\begin{split}
D_G' & = \frac{p_{r}(x)}{D(x)} + \frac{p_{g}(x)}{(1-D(x))}  \\
D_G^{\star} & = \frac{p_{r}(x)}{p_{r}(x)+p_{g}(x)}
\end{split}
.\end{equation}

By plugging the result into the objective function we have: 
\begin{equation}
V(D^{\star}, G)= \int p_{r}(x) log(\frac{p_{r}(x)}{p_{r}(x)+p_{g}(x)}) + p_{g}(x) log(\frac{p_{g}(x)}{p_{r}(x)+p_{g}(x)}) dx
.\end{equation}

The general aim of the process is to have a Generator mimicking perfectly the true distribution $p_{r}(x)$. Let set then ${p_{r}(x)=p_{g}(x)}$. It implies that that $D_G^{\star}=\frac{1}{2}$. The objective function becomes: 

\begin{equation}V(D^{\star}, G)= \int p_{r}(x) log(\frac{1}{2}) + p_{g}(x) log(\frac{1}{2})dx.\end{equation}

It is easy to see that $V(D^{\star}, G)= -2log(2)$ but we would like to do more. The aim is to prove that this minimum can be reached for only 
one G. 

\begin{equation}
\begin{split}
V(D^{\star}, G)
 & = \int p_{r}(x) log(\frac{p_{r}(x)}{p_{r}(x)+p_{g}(x)}) + p_{g}(x) log(\frac{p_{g}(x)}{p_{r}(x)+p_{g}(x)})dx\\
& = \int (log(2)-log(2))p_{r}(x) + p_{r}(x) log(\frac{p_{r}(x)}{p_{r}(x)+p_{g}(x)}) + \\ & (log(2)-log(2))p_{g}(x)  + p_{g}(x) log(\frac{p_{g}(x)}{p_{r}(x)+p_{g}(x)})dx \\
& = -log(2 \int {p_{r}(x)+p_{g}(x)} dx) + \int p_{r}(x)log(2){} + \\ & log(\frac{p_{r}(x)}{p_{r}(x)+p_{g}(x)})
 p_{g}(x)log(2)  + log(\frac{p_{g}(x)}{p_{r}(x)+p_{g}(x)})dx \\
& = -2log(2) + \int p_{r}(x)(\frac{p_{r}(x)}{(p_{r}(x)+p_{g}(x))/2})dx + 
\int p_{g}(x)\frac{p_{g}(x)}{(p_{r}(x)+p_{g}(x))/2}dx
\end{split}
\end{equation}

By definition of the Kullback-Leibler divergence \footnote{We will come back later on the Kullback-Leibler and Jenson-Shannon Divergences}, we have: 
\begin{equation}
V(D^{\star}, G) = -2log(2) + D_{KL}(p_{r}(x)\vert \frac{(p_{r}(x)+p_{g}(x))}{2}) \\ + D_{KL}(p_{g}(x)\vert  \frac{(p_{r}(x)+p_{g}(x))}{2})
.\end{equation}
The Kullback-Leibler divergence is non-negative so we confirm that $-2log(2)$ is a global minimum. 

And by definition of the the Jenson-Shannon divergence, we obtain: 

\begin{equation}
V(D^{\star}, G) = -2log(2) + D_{JS}(p_{r}(x)\vert p_{g}(x)). 
.\end{equation}

The Jenson-Shannon divergence is $0$ if, and only if, $p_{r}(x)= p_{g}(x)$). 

\subsection{Vanishing Gradient}
However, as mentioned in the original paper, the original loss function may not provide a gradient large enough for $G$ to learn. Indeed, 'Early in learning, when $G$ is poor, $D$ can reject samples with high confidence because they are clearly different from the training data. In this case,  $log(1-D(G(z)))$ saturates.' \citep[p. 3]{goodfellow2014a} Since it may not be obvious and it is not shown in the original paper, the proof is as follow:

\begin{equation}
\begin{split}
\nabla_{\theta} L_G &= \nabla_{\theta} \mathbb{E}_{z \sim p_z(z)}log(1-D(G_{\theta}(z))) \\
&=- \mathbb{E}_{z \sim p_z(z)} \frac{1}{(1-D(G_{\theta}(z)))} \frac{\partial D(G_{\theta}(z))}{\partial G_{\theta}(z)}  \frac{\partial G_{\theta}(z)}{\partial \theta} \\
&= \mathbb{E}_{x \sim p_g(x)} \frac{\nabla_x D(x) \nabla_{\theta} G_{\theta}(z)}{D(x)-1}  
.\end{split}
\end{equation}

If $D$ is a perfect Discriminator $D^{\star}$ then  $D^{\star}(x)|x \sim p_g(x)=0$, so that:
\begin{equation} \lim_{D \to D^{\star}} D(x) = 0 \end{equation} 
\begin{equation} \lim_{D \to D^{\star}} \nabla_\theta D(x) = 0 .\end{equation}

To prevent this issue, the authors propose to maximize $log(D(G_\theta(z)))$ instead of minimizing $log(1-D(G_{\theta}(z)))$.

\subsection{The Generator} 
The second part of the proposition shows that $p_g$, the distribution implicitly defined by $G$ converges to $p_{real}$. As $V(p_g, D)$ is convex in $p_g$ with a unique global optima, with sufficiently small updates, $p_g$ converges to $p_{real}$. 

However, in practice, $G$ represents only a limited family of functions $p_g$, the set of the functions $G(\theta)$. Consequently, we are optimizing a limited family of functions through $G(z, \theta)$. Moreover, we are not optimizing $p_g$ but $\theta_g$. As $G(z, \theta)$ is not a convex function so that we have no guarantee to reach the global minimum. These limitations contribute to making GANs hard to train and unstable. 

To give a clearer idea of how GANs works in practice, we present a pseudo-code version of the algorithm. 

\vspace{\baselineskip}
\begin{algorithm}[H]
\SetAlgoLined
\SetKwInOut{Required}{Required}
\caption{Standard Generative Adversarial Networks Algorithm}
\Required{$m$ the minibatch size. A prior function generating $z$. The hyperparameters $d_{steps}$ and $g_{steps}$. An algorithm for gradient ascent. The number of iterations.}
 \For{number of iterations}{
    \For{$d_{steps}$}{
      Sample ${z^{(1)},...,z^{(m)}}$ from noise prior $p_g(z)$\;
      Sample minibatch of examples ${x^{(1)},...,x^{(m)}}$ from data $p_{data}(x)$\;
      Update $\theta_D$ by ascending:
      \begin{equation}\nabla_{\theta_D}  \sum_{i=1}^m log( D(x^{(i)}))+ log (1-D(G(z^{(i)})))\end{equation}
    }
    \For{$g_{steps}$}{
      Sample ${z^{(1)},...,z^{(m)}}$ from noise prior $p_g(z)$\;
      Update $\theta_G$ by ascending the non-saturating function:
      \begin{equation}\nabla_{\theta_G}  \sum_{i=1}^m log (D(G(z^{(i)})))\end{equation}
      }
} 
\end{algorithm}

\subsection{Instability and Mode collapse}
The main problem with GANs is that they are known to be hard to train. The problems appear when one of the networks is too strong for the other. 
If the Discriminator becomes too strong, meaning its classifying correctly with high confidence all the observations, the Generator would not be able to learn anything. Indeed, if $D(G(z))$ is close to one, $log(1-D(G(z)$ saturates and its gradient vanish. As seen above, the authors of the original paper propose to prevent it by maximizing log(D(G(z))) instead of minimizing $log(1-D(G(z)))$. By changing the sign of the new expression the objective becomes again a minimization problem. The new loss function is then: 
\begin{equation}\max\limits_G \min\limits_D V(D, G) = \mathbb{E}_{x\sim p_{r}(x)} [log(D(x))] - \mathbb{E}_{z\sim p_{z}(z)}[log (D(G(z)))]
.\end{equation}

The strength of the Generator is also a problem. The reason is the so-called mode collapse. If $G$  is learning too fast, it will learn to produce data where the Discriminator is the most likely to be fooled, for instance, the mode of the objective function for a fixed $D$. The Discriminator will then progressively learn to classify this mode as fake. When $D$ is good enough at predicting fake for this mode, the Generator will find another weakness in the Discriminator and collapse to this new mode, until the Discriminator adjusts and predicts it as fake. The Generator comes back to the initial state and the same process occurs again. Even if a full mode collapse is rare, partial ones, happening only on a part of the distribution, are very frequent.

\section{GANs variations}

The address of this issue has driven to the production of many propositions to fix it. The field is new and fast-growing. In addition to the Standard GAN (SGAN), which we just presented, there exists a long list of alternative loss functions each of them bringing a new name for the GAN. Therefore, it is hard to be exhaustive because there are so much of them. We will present here only the versions we have worked with: the Wasserstein GAN (WGAN) and its variation with Gradient Penalization (WGAN-GP) and the  Relativistic Standard GAN, and its variation the Relativistic Averadged Standard GAN.

\subsection{Wasserstein GAN (WGAN)}

The Wasserstein GAN (WGAN) is often seen as the state-of-the-art. The solution has been presented by \citep{Arjovsky2017} in 2017. They propose to change the loss function. For them, the problem with the standard GAN's loss function is that the gradient of $log(D(p_g(x)))$ has a high variance when $p_g(x)$ is close to 1. 
The new proposed loss function lay on the Wasserstein distance also called Earth-Mover distance. The idea is that the distance between two distributions is the minimum distance of density (thought as earth) to move in order that the two distribution matches. Formally the Wasserstein distance is written: 
$$ W(p, q) = \inf_{\gamma \in \Pi(p, q)} \iint\limits_{x,y} \lVert x - y \lVert \gamma (x,y) \, \mathrm{d} x \, \mathrm{d} y  = \inf_{\gamma \in \Pi(p, q)} \mathbb{E}_{(x, y)\sim\gamma}(\left\lVert x-y \right\rVert)  $$

where $\Pi(p, q)$ denotes the set of all joint distribution $\gamma(x, y)$ whose marginals are respectively $p$ and $q$. The above formulation is intractable. 
The authors of the original paper propose then a transformation using the Kantorovich-Rubinstein Duality. The detail of the transformation is beyond the scope of this work. However, for details about it, Vincent Herrman wrote a convincing explanation \citep{Herrmann}. The result is that we can also express the former formulation as follow: 

$$W(p_r, p_g) = \frac{1}{K} \sup_{\| f \|_L \leq K} \mathbb{E}_{x \sim p}(f(x)) - \mathbb{E}_{x \sim q}(f(x))$$

The new formulation makes appear the condition $\| f \|_L \leq K$, meaning that $f$ should to be K-Lipschitz continuous. A real function $f: \mathbb{R} \rightarrow \mathbb{R}$ is said K-Lipschitz if there is a real constant $K$ such that 

\begin{equation} \lvert f(x_1) - f(x_2) \rvert \leq K \lvert x_1 - x_2 \rvert \forall x_1, x_2 \in \mathbb{R} .\end{equation}.

In other words, for differentiable functions, the slope should not be larger than $k$.

The idea is to make the Discriminator such that it belong to the family of K-Lipschitz continuous functions $f_{\theta}$. 
It would be then possible to rewrite the loss function of the Discriminator as a Wasserstein distance. Assuming that $f_{\theta}$ is K-Lipschitz, the loss function becomes: 
\begin{equation}L(p_r, p_g) = W(p_r, p_g) = \max_{\theta} \mathbb{E}_{x \sim p_r}(f_\theta(x)) - \mathbb{E}_{z \sim p_g(z)}(f_\theta(G(z))).\end{equation}
So, the Discriminator is not a classifier anymore. It is rather a function required to compute Wassstein distance. To emphasize this change the Discriminator is renamed \textit{Critic}. 
The issue is now to keep $f_\theta$ being a K-Lipschitz continuous function so that the above developments hold. The trick used is quite simple. It consists of clamping the weight after each update in order that the parameter space being compact. Thus, $f_\theta$ is bounded such that the K-Lipschitz continuity is preserved. The weights are clamp in an interval $[-c,c]$, where $c$ is a small value. The authors recommend $c$ being of the order of $10^{-2}$. According to the authors themselves, 'weight clipping is a terrible way to enforce a Lipschitz constraint. (...) However, we do leave the topic of enforcing Lipschitz constraints in a neural network setting for further investigation, and we actively encourage interested researchers to improve on this method.' \citep{Arjovsky2017}(p.7). 

The authors of \citep{Gulrajani2017} have followed this recommendation and they proposed a penalization of the gradient to ensure that the Lipschitz constraint hold. Instead of clipping the weight, they propose to penalize the norm of the gradient take values far from one. 
To do that they defined $\hat{x}$ as a value uniformly sampled along straight lines between pairs of points themselves sampled from the real distribution $p(x_r)$ and the generated distribution $p(x_r)$, which implicitly define $p_{\hat{x}}$ such that $\hat{x}\sim p_{\hat{x}}$. The new loss function with gradient penalization becomes:

 \begin{equation} L= \mathbb{E}_{x \sim p_r}(f_\theta(x)) - \mathbb{E}_{z \sim p_g(z)}(f_\theta(G(z)))\ + \lambda \mathbb{E}_{\hat{x}\sim p_{\hat{x}}}[ (||\nabla_{\hat{x}}D(\hat{x})||_2-1)^2 ], \end{equation} where $\lambda$ is a new hyper parameter controlling the strength of the constraint. The authors of the original paper, dealing with high dimensional input, propose to use $\lambda=10$. Our experiments showed that this value has to be tuned for smaller input \footnote{This is typically the process that seems of little importance but which is in practice very time consuming because the algorithm does not work with the good value. In particular,  the first time we implemented a GAN-GP, it is hard to say, where the problem comes from. }. We used $\lambda=0.1$. 
 
WGANs solve the problem of vanishing or high variance gradients. They offer also a convincing way of computing distances between distributions for the kind of issue they deal with. The theory is convincing. The problem is that it is not clear that they perform better than classic GAN. November 2017, a team from Google Brain published a paper called 'Are GANs created equal?', comparing between 8 different GANs, including the 2 original versions (with and without the non-saturating transformation), and the two versions of the WGAN presented here. They conclude: 
'We did not find evidence that any of the tested algorithms consistently outperform the non-saturating GAN introduced in [the original paper]'. 

\subsection{RSGAN and RaSGAN}
\citet{Jolicoeur-Martineau2018} proposed a new class of GANs, called relativistic GANs. The idea comes from the argument 'that the key missing property of SGAN is that the probability of real data being real $(D(x_r))$ should decrease as the probability of fake data being real $(D(x_f))$ increase' (p.3). This main argument is based on the fact that the Discriminator can use the \textit{a priori} knowledge that half of the examples are real and half are fake. 

Just like for the WGAN, we define the non-transformed output of the Discriminator as $C(x)$. The standard Discriminator is then defined as $D(x)=\sigma(C(x))$. We defined then $\Tilde{x}$ as a pair of a real and fake data $(x_r, x_f)$. A relativistic can be easily defined as $D(x)=\sigma(C(x_r)-C(x_f))$, where $\sigma$ is the sigmoid function. Following \citet[p.5]{Jolicoeur-Martineau2018}, 'we can interpret this modification in the following way: the Discriminator estimates the probability that the given real data is more realistic than a randomly sampled fake data'. The opposite relation $D_{rev}(x)=\sigma(C(x_f)-C(x_r))$ doesn't need to be include in the loss function since $1-D_{rev}(x)=D(x)$. The loss function of a relativistic GAN can be written as follow:
\begin{equation} L_D = -\mathbb{E}_{(x_r, x_f) \sim \mathbb{P}, \mathbb{Q}} [log(\sigma(C(x_r)-C(x_f)))]\end{equation} 
\begin{equation} L_G = -\mathbb{E}_{(x_r, x_f) \sim \mathbb{P}, \mathbb{Q}} [log(\sigma(C(x_f)-C(x_r)))].\end{equation}

The principle of the relativistic GAN can be applied to a broad family of GANs by replacing the sigmoid function by any other function. 

The Relativistic average standard RaSGAN is very similar to the RSGAN except that the non-transformed output of the Discriminator is not compared to a single data of the other type but to the batch average. We can interpret it as 'the probability that the input data is more realistic than a randomly sampled data of the opposing type (fake if the input is real or real if the input is fake)' \citet[p.6]{Jolicoeur-Martineau2018}. The loss function for the Discriminator and the Generator have the same expression. It is a bit longer than the one of the RSGAN since the two parts of the expression have to be included:

\begin{equation} L = -\mathbb{E}_{x_r \sim \mathbb{P}}[log(\sigma(C(x_r)-\mathbb{E}_{x_f \sim \mathbb{Q}}C(x_f)))] -\mathbb{E}_{x_f \sim \mathbb{Q}}[log(1-\sigma(C(x_f)-\mathbb{E}_{x_r \sim \mathbb{P}}C(x_r)))].\end{equation} 

There exists a broad class of GANs with different objective functions. We presented here only some of the most popular but we also ignored others. The choice depends on the popularity and the presumed efficiency but we have to admit the selection is also arbitrary. Moreover, as the number of papers published about GAN is high, it is hard to stay up to date. We will come back later on the specific architecture and hyperparameters used in the experiment section.

\section{Conditional GAN}
 We remind that GAN aims to "learn" the distribution of a probabilistic model from a sample and consequently sampling from this approximation. Therefore, it is also possible to build a conditional version of it. Instead of "learning" $p(X)$, the generative model will learn $p(X|Y)$. 
 
 The idea has been introduced shortly after the original GAN by Mirza and Osindero \citep{mirza_conditional_2014}.  The task looks even easier in the sense that both Generator and Discriminator will have more information. In practice, the change is straight forward. Instead of giving just noise, we give the joint distribution $p(Y, Z)$ to the Generator and the joint distribution $p(Y, X)$ to the Discriminator. The $Y$'s come from the dataset and are randomly shuffled to form batches of $Y$'s. The Generator will "learn" to produce the joint distribution $p(X, Y)$ and the Discriminator will discriminate real and fake knowing the couple $(Y, X)$. In other words, the Discriminator is also distinguishing real and fake using the joint distribution. 
 
 In detail, the Discriminator will learn to classify the probability distribution of $p(X, Y)$. So that, the classification between generated $X$'s and real $X$'s will depend on $Y$. In return, the Generator has to produce a realistic joint distribution $p(X, Y)$.
 
 Formally, the transformation is straight forward: 
 
 \begin{equation} \max\limits_D V(D, G) = \mathbb{E}_{x\sim p_{r}(x)} [log(D(x, y))] + \mathbb{E}_{z\sim p_{z}(z)} [log (1-D(G(z, y)))] .\end{equation}
 
 The demonstration is exactly equivalent to the case without condition.
 
\begin{figure}[ht!]
\centering
\includegraphics[scale=0.3]{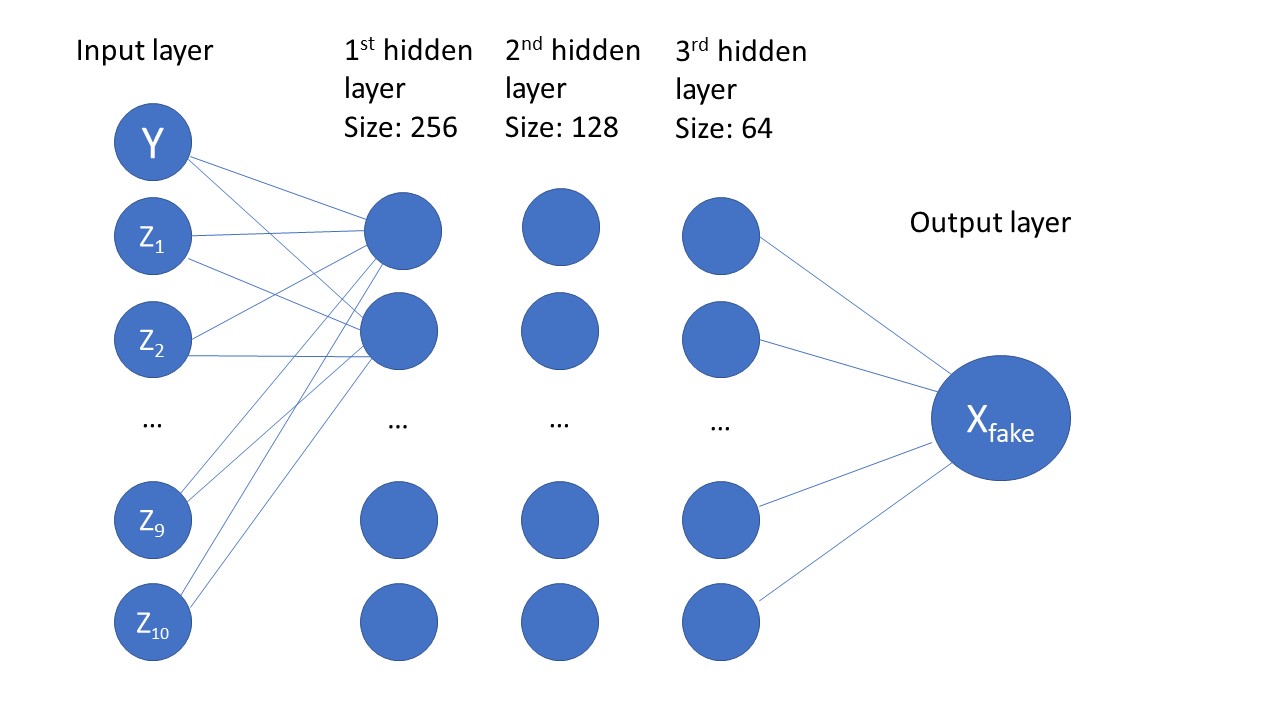}
\caption{The Conditional Generator}
\label{Gene}
\end{figure}

\begin{figure}[ht!]
\centering
\includegraphics[scale=0.3]{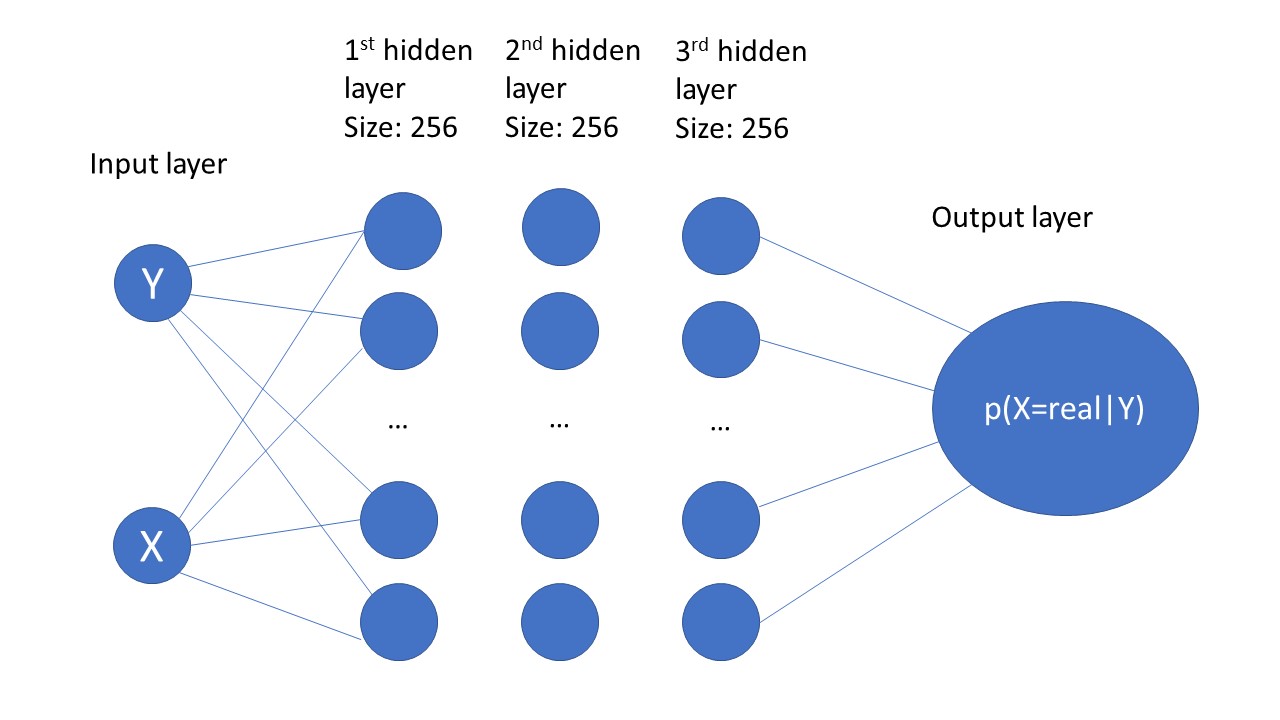}
\caption{The Conditional Discriminator}
\label{Disc}
\end{figure}

 The Generator takes $Z$ and $Y$ as inputs and produces samples simulating For more clarity, we propose graphical representation of the  Generator and the Discriminator we are going to use respectively in Figure \ref{Gene} and Figure \ref{Disc}. 
 
 \subsection{Conditional GAN for regression}
 
 As introduced in the first chapter, this conditional distribution corresponds to the probability distribution for a new observation in a regression. Therefore, this conditional distribution can be used as the predictive distribution of a regression. However, the explicit probability distribution remains unknown.  
  
 Once the Conditional GAN has converge, it produces sample for the conditional distribution such that $p_{g}(X|Y) \approx p_{r}(X|Y)$. In other words, the Generator acts like an implicit distribution function. It generates samples from the approximate distribution. So, the empirical distribution can be used to estimate properties of the true conditional distribution.
 
 Practically, we will use the trained Generator as follow. It will take the conditional value (the new observation), say $Y$, for which we would like an estimate and the random noise using the normal distribution $Z \sim \mathcal{N}(0, I_n)$. By sampling a large number of $Z$ and keeping fixed the condition, we obtain a large sample of data following the approximated conditional distribution.
 
 From this sample, we can theoretically estimate any kind of statistics about $p_{g}(X|Y)$, including the moments of arbitrary large order and all the quantile. Indeed, GANs can be seen as a class of Monte Carlo sampler. Indeed, empirically, GANs are related to Markov process but the mathematical relation between them is beyond the scope of this work. 
 
 It is interesting to notice that GANs can produce conditional distributions in both directions, i.e. $p_{g}(X|Y)$ as well as $p_{g}(Y|X)$. In other words, conditional GANs can produce predictions for classical or reversed regression. Moreover, the dimensions of $X$ and $Y$ have no restriction in neither direction. In principle, the use of conditional GANs for regression task is very flexible. 
 
  We call this use of conditional GANs in a regression purpose \textit{adversarial regression}. The next part of this work consists of testing how well the adversarial regression performs in approximating statistics about the true density for a new observation.

\chapter{Experiments, metrics and hyper-parameters}

In this short chapter, we would like to present first the kind of regression experiments that we are going to test, second, the metrics that we will use and, third, the settings used for all the hyperparameters. 

\section{The experiments}

The aim of this work consists of testing different kinds of GANs and parameters for regression tasks. We define three simple ‘true’ models to this end. We sample from these models a number $n$ of events ($n$ is part of the factors tested) and we run the adversarial regression on the samples. We approximate a new observation using the conditional distribution for a given value thanks to the conditional GAN. We compare it to the true distribution. We choose three bivariate scenarios, which are much easier to analyze. In particular, the prediction distribution depends only on one variable. Of course, this is a strong limitation of these experiments. Further experiments in higher dimensions have to be led to confirm the result we have found.  

Let have a look first at the three models. The first one corresponds to the classical linear model with normal errors :
\begin{equation}Y = aX+b + \epsilon,  \epsilon \sim N(0, 0.0025).\end{equation} The second model is still linear but with heteroscedasticity: \begin{equation}Y = aX+b + \epsilon,   \epsilon \sim N(0, x^2*0.01). \end{equation}
The third model is a mix of a linear trend, a sinusoidal oscillation and an homoscedastic error term. It is defined as follow : 
\begin{equation}Y = aX+ c*sin(dx) + \epsilon, \epsilon \sim N(0, 0.0025).\end{equation}

\begin{figure}[ht!]
\centering
\includegraphics[scale=0.4]{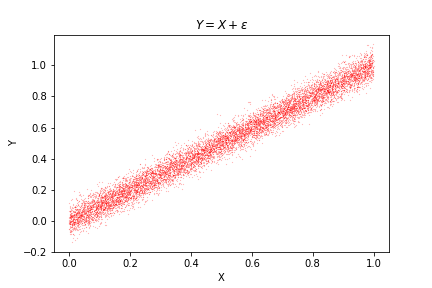}
\caption{10'000 realizations of model 1}
\label{fig:figure 3}
\end{figure}

\begin{figure}[ht!]
\centering
\includegraphics[scale=0.4]{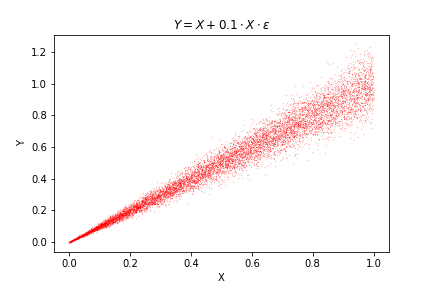}
\caption{10'000 realizations of model 2}
\label{fig:figure 4}
\end{figure}

\begin{figure}[ht!]
\centering
\includegraphics[scale=0.4]{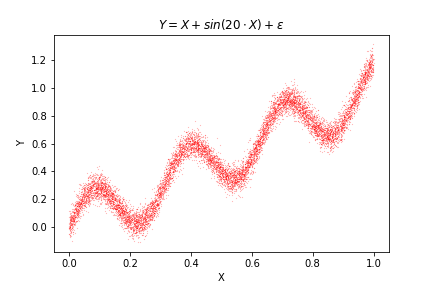}
\caption{10'000 realizations of model 3}
\label{fig:figure 5}
\end{figure}

$X$ is considered as the independent variable. We generated it uniformly on the interval $[0, 1]$.  For these experiments, we fixed $a=1$, $b=0$, $c=0.2$ and $d=20$. The figures \ref{fig:figure 3} to \ref{fig:figure 5} show 10'000 realizations of these models. 

On each of these models, we can perform two kinds of experiments. First, the classical regression version consisting of predicting $Y$ for a given $X$. But we can also think the inverse problem and be interested to find the distribution of $X$ given $Y$. It is known as inverse regression or calibration problem. We will present results for both cases. However, most of the experiments where run for inverse regression because the inverse regression problem proposes more various situation, with a broader range of complexity. Indeed, the first experiment is exactly identical in both directions, the normal distribution being symmetric around zero. In the two other cases, however, $p(X|Y)$ has a different and often unknown distribution so that the distribution cannot be found analytically. To prevent this issue, the true distribution $p(X|Y)$ will be estimated by sampling and slicing. Practically, we have sampled $10^7$ events from the true distribution and estimated the true distribution $p(X|Y)$ with all the event contained in the slice $Y \pm 0.01$. We also stored the number of events from the true distribution to generate the same number of 'fake' data.

\begin{figure}[H]
\centering
\def\tabularxcolumn#1{m{#1}}

\subfloat[$y=0.1$]{\includegraphics[width = 3cm]{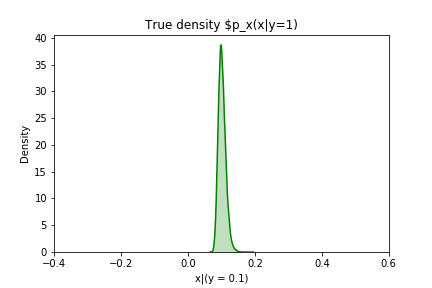}} 
\subfloat[$y=0.4$]{\includegraphics[width = 3cm]{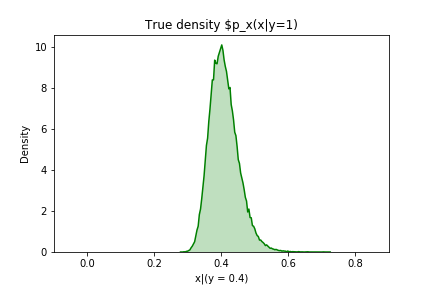}}
\subfloat[$y=0.7$]{\includegraphics[width = 3cm]{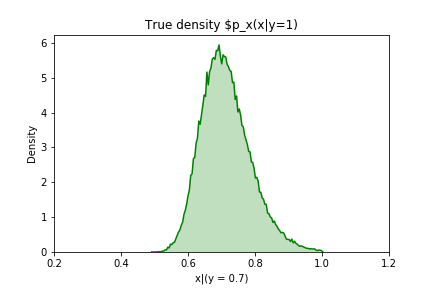}} 
\subfloat[$y=1.0$]{\includegraphics[width = 3cm]{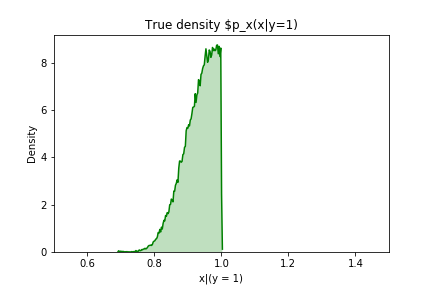}}

\caption{Model 2: True densities for chosen values of $y$}
\label{Truedens2}
\end{figure}

We have arbitrary selected four values to condition on. These values are 0.1, 0.4, 0.7, 1. They are the same in all experiments whether we condition on $x$ or $y$. These values ensure to have measures for a broad range of the distribution. In particular, they include various variances in case of heteroscedasticity and produce diverse shapes for the inverse regression of the third experiment. Because the values $Y=1$ and $X=1$ are on the edge of the distribution, we note that it produces particular situations. When looking for the prediction $p(Y|X=1)$, the prediction is harder because there is no bigger $x$'s. In the inverse case, looking for $p(X|Y=1)$, the true distribution fall suddenly to zero, except in the third experiment. For instance, Figures \ref{Truedens2} and \ref{Truedens2} shows the true conditional distributions of $p(X|Y)$ for the experiment 2 and 3. As explained, the true densities plotted are estimated by sampling and slicing around the selected $y$'s.

\begin{figure}[H]
\centering
\def\tabularxcolumn#1{m{#1}}

\subfloat[$y=0.1$]{\includegraphics[width = 3cm]{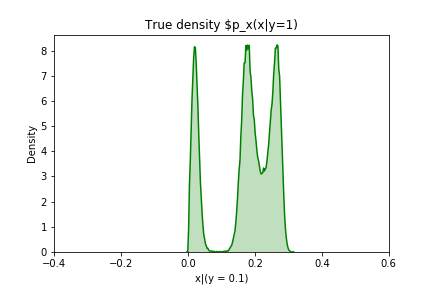}} 
\subfloat[$y=0.4$]{\includegraphics[width = 3cm]{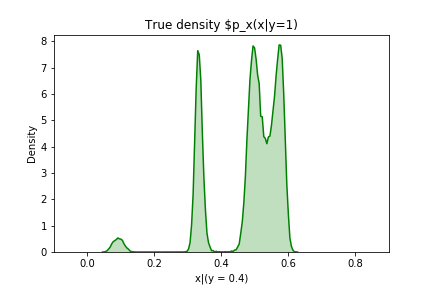}}
\subfloat[$y=0.7$]{\includegraphics[width = 3cm]{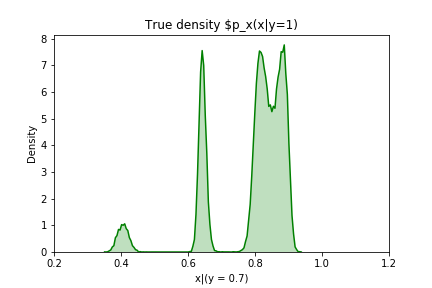}} 
\subfloat[$y=1.0$]{\includegraphics[width = 3cm]{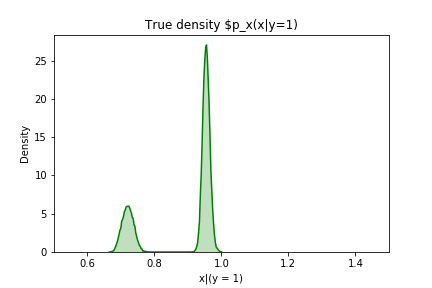}}

\caption{Model 3: True densities for chose values of $y$}
\label{Truedens}
\end{figure}

\section{The metrics}
Measuring the distance between two samples can be complicated in a high dimensional space. Fortunately, there are plenty of different measures in the uni-dimensional case.
The distance between the real and the generated means is a rough and obvious measure. As we are interested in the whole distribution, it is not enough. The distance between the next moments, variance, skewness and kurtosis refine the comparison. We are going to use these distances since they are common and frequently used. They indicate if the approximation is roughly correct. However, they are quite complicated to interpret. For instance, we are annoyed to give interpretation when the higher moments are well estimated but not the lower one. We would prefer a more global measure. 
Fortunately, there are also good metrics giving a general distance or pseudo-distance between samples. We have considered three of them. The first one is the Kolmogorov-Smirnov distance. The second is the KL divergence. The third one is the JS divergence. Let have a look in detail on these three measures. 

\subsection{The KS distance}
The Kolmogorov-Smirnov statistics or distance measure the supremum of the absolute value of the difference between the cumulative distributions functions. Formally, we write:
\begin{equation} D_{KS_{n,m}} = \sup_{x} |F_{n, P}-F_{m, Q}|. \end{equation}
The Kolmogorov-Smirnov statistics is a non-parametric statistic, known above all for its use in the Kolmogorov-Smirnov test, which is a test of goodness of the fit.  

\subsection{The KL divergence}
The Kullback-Leibler Divergence is a measure of how close a distribution is from another. It is defined as follow:
\begin{equation} D_{KL}(P||Q) = \int_{-\infty}^{\infty} p(x) ln(\frac{p(x)}{q(x)})dx.\end{equation}

However, in our case, we do not have a closed expression neither for $p(x)$ nor $q(x)$. To prevent this issue, we are going to use the discretized version of the divergence to approximate the true divergence. In practice, we are going to divide the output space between $-0.5$ and $1.5$ in 200 bins \footnote{We know that the distribution outside this interval is almost zero.}, approximate the continuous distribution through the probability mass function of those 200 bins and compute the discrete Kullback-Leibler Divergence, defined as follow:

\begin{equation} D_{KL}(P||Q) = \sum_{x \in \mathcal{X}} P(x) ln(\frac{P(x)}{Q(x)}).\end{equation}

We remark that the Kullbach-Leibler Divergence take value from zero to infinity. We note also that it is not properly a distance since it is asymmetric: $D_{KL}(P||Q) \neq D_{KL}(Q||P)$.

\subsection{The JS divergence}
The Jenson-Shannon Divergence can be interpreted as the mean of the Kullbach-Leibler divergence between each distribution and the mean between them. The definition is:
\begin{equation}D_{JS}=\frac{1}{2}D_{KL}(P||M) + \frac{1}{2}D_{KL}(Q||M),\end{equation} where $M= \frac{1}{2}(P+Q)$
It is easy to show that the range of the Jenson-Shannon Divergence is contained between $0$ and $ln(2)$. The Jenson-Shannon Divergence has the convenient property to be symmetric. Moreover, Jenson-Shannon Divergence is the metric that GAN aims to minimize.

We have led the experiments computing the three distances. In practice, the three distances give very close results. In most of the cases, the are very redundant. We have not seen an experiment where one of the metrics have brought significant additional information than the others. We will give the KS distance, the JS and the KL Divergence as well as the distance between means only in the first part of the next chapter. It will illustrate the high correlation between them. After that, we will only give the Jenson-Shannon (JS) Divergence.

\subsection{The variance of the measure}
\begin{figure}[ht!]
\centering
\includegraphics[scale=0.4]{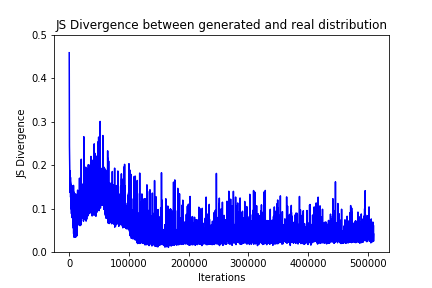}
\caption{Example: The evolution of JS Divergence during training}
\label{Train}
\end{figure}

We have some metrics to compare a sample coming from the true distribution and a sample coming from the generated one but the issue about the metric to measure the performance of GANs is not done yet. GANs are iterative processes. Therefore, the Generator is updated at every iteration, so that at every iteration the function $G(Z)$ is different and yield different samples. Figure \ref{Train} shows such an evolution of the distance among iterations. As we can see, the quality of the sample is quite noisy between iterations. Consequently, a single measure of the distance at the $i^{th}$ iteration is not relevant. 

To address this issue, we choose to measure the distance every 100 iterations. However, we will gives some distances only every 10'000 iterations. The 100 distances measured between the 10k updates will be used to compute statistics about the distance. For instance, instead of giving the JS Divergence after 10k iterations, which is a noisy measure, we will give the average distance of the 100 measures between 10'000 and 19'900. It is then possible to analyze a lot of interesting statistics about the evolution of the distance: the mean, the variance, the minimum, the maximum. Ideally, we want the distance not only being minimal but also a small variance, and a low maximum. In practice, we are going to use the mean, the third quarter and the maximum of this 100 iterations. 

In most cases, however, the variance is similar for the same type of GAN. Therefore, we will give only the mean for conciseness. The third quartile and the maximum are interesting mostly while comparing the different types of GANs. We will use it only on that purpose.

\section{GANs and hyperparameters}
The following experiments fixed some hyperparameters. We make vary others to test is. However, some values are taken as the values by default, with which others hyperparameters are tested.  

First of all the architectures of all the GAN are fixed. They consist of two neural networks with 4 layers. Each hidden layer is a fully connected layer. The Generator takes as input the two variables $x$ and $y$ as well as the noise vector. The following layers are respectively of size  256, 128, 64 and 1 for the output layer.

The Discriminator and the Critic take the two variables as input. They have three hidden layers of size 256 and an output of dimension one. The only difference between the Discriminator and the Critic is that the Discriminator transform the output contains a  sigmoid transformation while the Critic is simply a linear combination of the outputs of the previous layer. 

Except for the output layer, the non-linear functions are Relu's for the Generator and Leaky Relu with parameter $\alpha=0.2$ for the Critic or Discriminator. 
Finally, for all the networks, we used an Adam optimizer with betas equal to 0 and 0.9, which is equivalent to use a stochastic gradient descent with momentum. The learning rate has been set at $10^{-4}$ for the SGAN and $10^{-5}$ for the other types of GANs. The Discriminator or the Critic is updated 5 times before an update of the Generator, except for the RSGAN where Critic and Generator are updated alternatively. 
These values have been found the hyperparameter search and based on shared good practices as they have been observed by the community.  For instance, the optimizer choice and its values for the parameters are also the ones used in \citep{Gulrajani2017}.  

Besides of these fixed values, we defined some default values. The default number of data in the sample is 10'000. The default dimension size for the noise value is 10. The batch size by default change depending on the type of GAN. There are respectively of 2000, 2000 and 500 for the SGAN, the WGAN-GP, and the RSGAN. These values are assumed to values providing good results accordingly to preliminary experiments. Since WGAN-clip and the RaSGAN provided comparatively poor results, we decided to exclude them of the results for more concision and clarity.

Table \ref{table:hyper} summarizes the hyperparameters, their aims and their values. We present in the same table the models and the size of the dataset even if they are not properly speaking hyperparameters.

\begin{table}[ht]
\begin{tabularx}{\textwidth}{|l |X |X|} 

\hline
 hyperparameter & Description & Values  \\
 \hline \hline
 Generator architecture & Number of layers and nodes & 4 layers with respectively 256, 128, 64, 1\footnote{5 in the last case presented in chapter 4. In general, it depends on the dataset.} nodes. Fixed. \\  \hline
 Discriminator arch. & Number of layers and nodes & 4 layers with respectively 256, 256, 256, 1 nodes. Fixed. \\ \hline
 Activation function & Function apply at each node after  the linear combination & Relu's for G leaky relu's for D. Fixed \\
 \hline
 Optimizer & Function used to updates the parameters & Stochastic gradient descent  with parameter=0.9.  Fixed \\
 \hline
 Learning rate & Parameter defining the size of  the step at each update & for SGAN: $10^{-4}$ else: $10^{-4}$.  Fixed \\
 \hline
 $d$ Discriminator updates. & Number of times the Discriminator is updates before the Generator also is & for RSGAN: 1 else: 5.  Fixed \\
 \hline
 Dimensionality of $Z$ & Size of the noise vector  used as input of G & Tested: 1, 2, 3, 5, 10, 20.  Default: 10 \\
 \hline
 Batch size & Number of event used for one update & Tested: 20, 80, 250, 500, 1000, 2000. Default: 2000. (500 for RSGAN) \\
 \hline
 Type of GAN & Type of the loss function giving  the name to the GAN  & Tested:SGAN, WGAN-GP, RSGAN  Default: SGAN \\
 \hline
 Models & Function used to generate the data & Tested: Model 1, 2 or 3, classical or inversed \\
 \hline
 Dataset size $n$ & Number of events contained in the dataset & Tested: 250, 500, 1'000, 2'500,  5'000, 10'000, 100'000 Default: 10'000 \\
 \hline
\end{tabularx}
\caption{Hyperparameters}
\label{table:hyper}
\end{table}

In this chapter, we have presented and justify the choice of the experiments, of the metrics and of the hyperparameters tested. In the next chapter, we are going to use them to assess the quality of various GANs and various hyperparameters.  

\chapter{Results}

We have trained many GANs of various types, with various hyperparameters on various experiments. We present in this chapter the main results of these experiments. In the first part, we are going to show how a Standard GAN can perform in classical and inverse regression tasks to have a better idea of the kind of results we can expect. In the second part, we are going to test several types of GANs with different hyperparameters. In the third part, we are going to see a few exploratory additional experiments and, on this basis, discuss the results.

\section{First overlook about adversarial regression}

This first part aims to verify that the adversarial regression works on the bivariate case and that we can produce accurate distributions of predictions. In this first part, we will use only Standard GANs (SGANs).

\subsection{Linear regression with Normal errors}

\begin{figure}[H]
\centering
\def\tabularxcolumn#1{m{#1}}
\subfloat[$x=0.1$]{\includegraphics[width = 3cm]{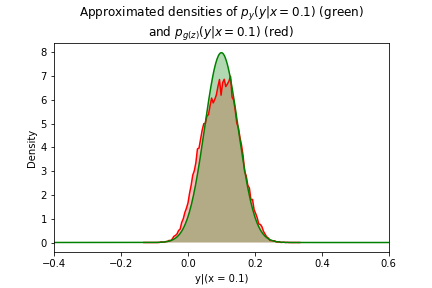}} 
\subfloat[$x=0.4$]{\includegraphics[width = 3cm]{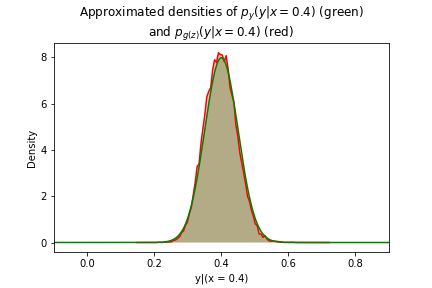}}
\subfloat[$x=0.7$]{\includegraphics[width = 3cm]{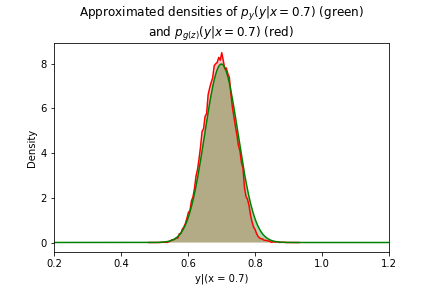}}
\subfloat[$x=1.0$]{\includegraphics[width = 3cm]{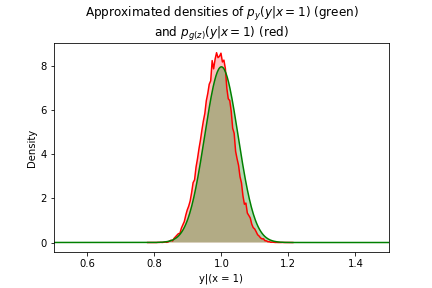}}
\caption{Model 1: True and generated densities}
\label{figure6}
\end{figure}

We start with the very basic linear regression problem with normal errors. It is a relatively easy task and we expect that our GANs will perform well. We tested it with a dataset of 10000 events. In Figure \ref{figure6}, the green curve is the true density, the red curve is the generated one. The 4 plots represent the approximate density of $y$ for the four selected given $x$'s. The generated density is an approximation based on a Gaussian kernel density estimate with 100'000 generated data and a bandwidth of 0.01. We can see that broadly the location and the scale of the distribution is well approximated.

\begin{figure}[H]
\centering
\def\tabularxcolumn#1{m{#1}}

\subfloat[$x=0.1$]{\includegraphics[width = 3cm]{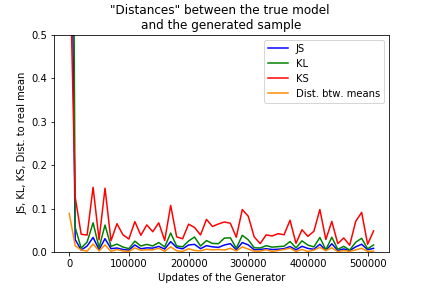}} 
\subfloat[$x=0.4$]{\includegraphics[width = 3cm]{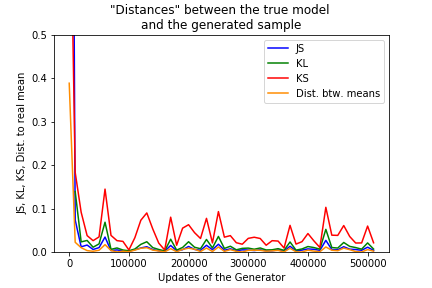}}
\subfloat[$x=0.7$]{\includegraphics[width = 3cm]{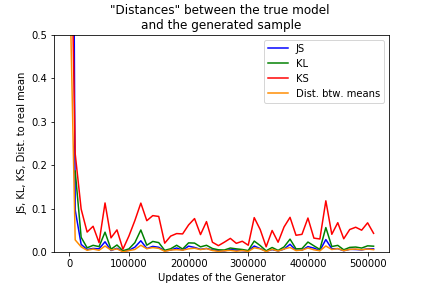}} 
\subfloat[$x=1.0$]{\includegraphics[width = 3cm]{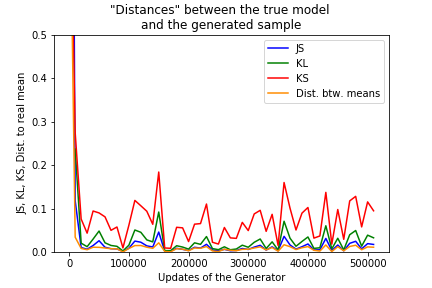}}

\caption{Model 1: Distances between true and generated distributions}
\label{figure7}
\end{figure}

Figure \ref{figure7} presents a batch of plots showing the evolution of the distances (or pseudo-distances) during the training. We plot the Jenson-Shannon divergence (JS), the Kullback-Leiber divergence (KL), the Kolmogorov-Smirnov distance (KS) and the distance between the means of the conditionally generated $x$ and the true expectation of Y.  
We observe that we reach quickly good values for all the four distances. For the four given $x$'s, the distances are not decreasing anymore after 20'000 updates of the Generator.
We observe that the different distances are, as expected, strongly but not perfectly correlated. It is a good sign that they measure the same general idea but with different methods. 
We observe that the distances are quite noisy. Once the training has converged to a minimum, the generate distribution oscillates around the true distribution.

\begin{figure}[H]
\centering
\def\tabularxcolumn#1{m{#1}}

\subfloat[Mean]{\includegraphics[width = 3cm]{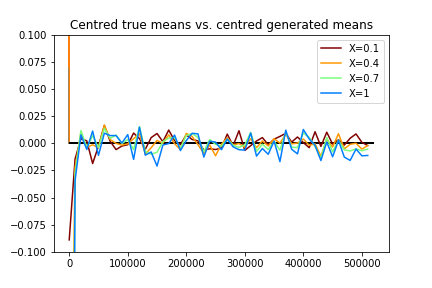}} 
\subfloat[Variance]{\includegraphics[width = 3cm]{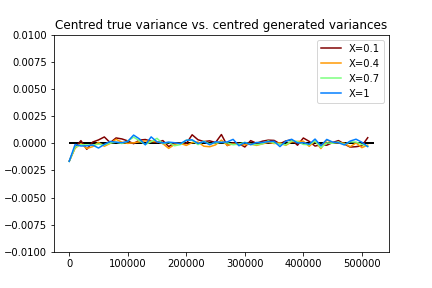}}
\subfloat[Skewness]{\includegraphics[width = 3cm]{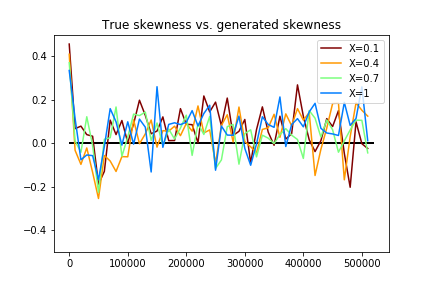}} 
\subfloat[Kurtosis]{\includegraphics[width = 3cm]{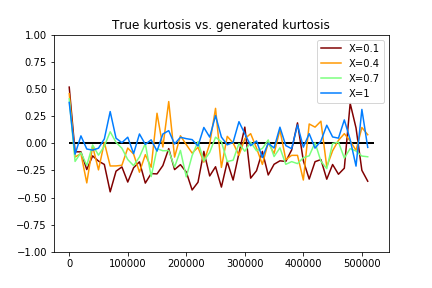}}

\caption{Model 1: Estimated moments from generated samples compared to true moments}
\label{figure8}
\end{figure}

We can also have a look at the difference between the theoretical true and the empirical generated four first moments. Again, Figure \ref{figure8} shows that the approximation is good enough and converge quickly around the true value, even though we can notice a small bias for the skewness and the kurtosis. Of course, all these estimates have a variance and, therefore, we can observe small fluctuations during training.

This was the easiest regression model possible. Is the adversarial regression able to produce as good results for more complex problems? Let's have a look at the three others cases: the one with non-normality of the error or heteroscedasticity, the one with non-linearity and the non-linear inverse regression problem.

\subsection{Linear regression with heteroscedasticity}

\begin{figure}[H]
\centering
\def\tabularxcolumn#1{m{#1}}

\subfloat[$x=0.1$]{\includegraphics[width = 3cm]{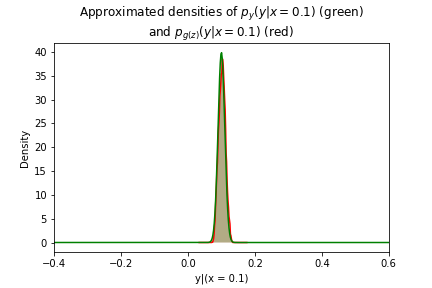}} 
\subfloat[$x=0.4$]{\includegraphics[width = 3cm]{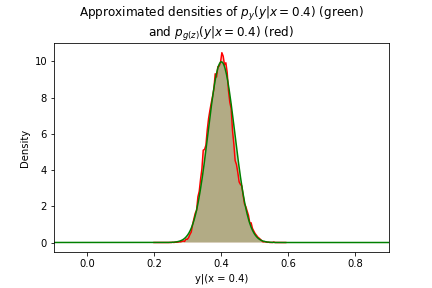}}
\subfloat[$x=0.7$]{\includegraphics[width = 3cm]{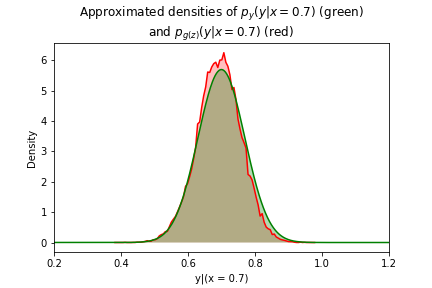}} 
\subfloat[$x=1.0$]{\includegraphics[width = 3cm]{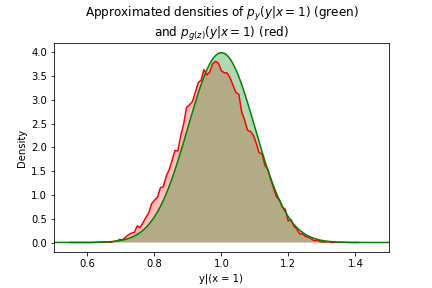}}

\caption{Model 2: True and generated densities}
\label{figure9}
\end{figure}

We present the same plots we have presented for the linear model with homoscedasticity but the model with heteroscedasticity. Again, the first check on the difference between the true density and the generated density, as shown in Figure \ref{figure9}, confirms that GAN can produce close approximations of the true density. 

\begin{figure}[H]
\centering
\def\tabularxcolumn#1{m{#1}}

\subfloat[$x=0.1$]{\includegraphics[width = 3cm]{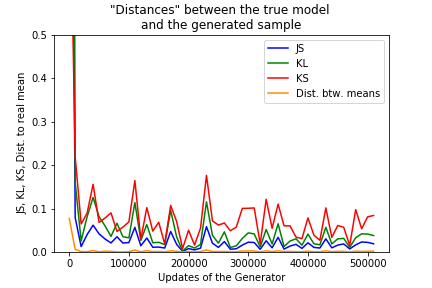}} 
\subfloat[$x=0.4$]{\includegraphics[width = 3cm]{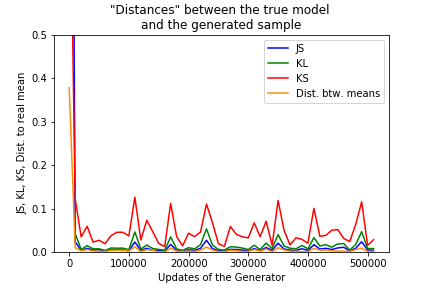}}
\subfloat[$x=0.7$]{\includegraphics[width = 3cm]{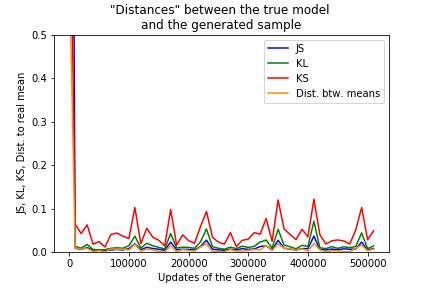}} 
\subfloat[$x=1.0$]{\includegraphics[width = 3cm]{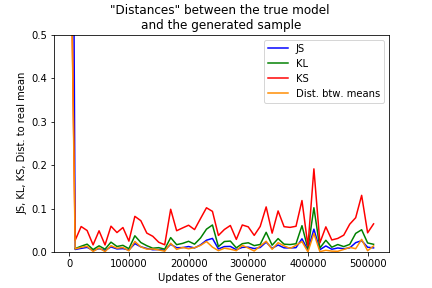}}

\caption{Model 2: Distances between true and generated distributions}
\label{figure10}
\end{figure}

\begin{figure}[H]
\centering
\def\tabularxcolumn#1{m{#1}}

\subfloat[Mean]{\includegraphics[width = 3cm]{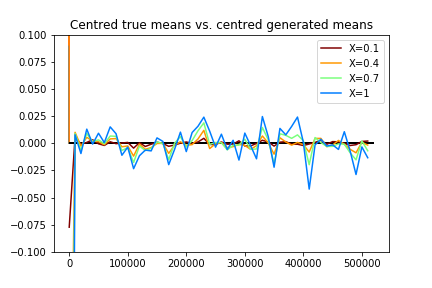}} 
\subfloat[Variance]{\includegraphics[width = 3cm]{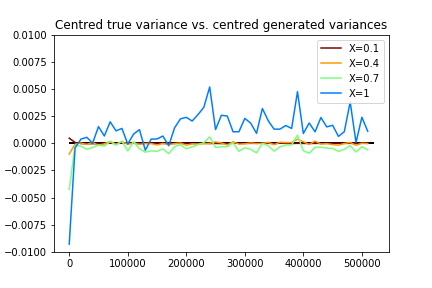}}
\subfloat[Skewness]{\includegraphics[width = 3cm]{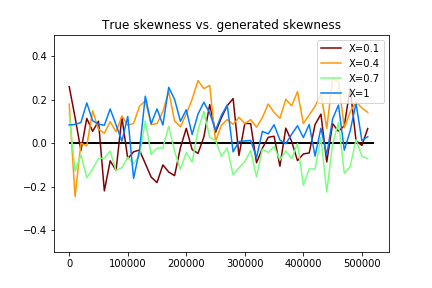}} 
\subfloat[Kurtosis]{\includegraphics[width = 3cm]{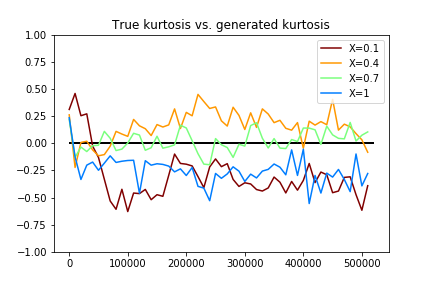}}

\caption{Model 2: Estimated moments from generated samples compared to true moments}
\label{figure11}
\end{figure}

The distances between distribution converge also quickly to a good approximation. Figure \ref{figure10} shows results very similar to the first case. However, the analysis of the distances between moments, plotted in Figure \ref{figure11} mitigate slightly the first visual assessment. We see that the approximation of the mean is worse than in the previous case. Especially, when the error term is large, it seems more complicated for the GAN to approximate well the expectation. The variance appears also harder to approximate when large. However, for the variance, this observation could have been expected. Indeed, since the variance is larger, it could have been expected that the error on the variance would have been larger too.
We can also observe that the estimates for the skewness and the kurtosis are also less accurate than in the first scenario, though their true value does not change. It is most likely the sign that this distribution represents a slightly more complex model and therefore is harder to learn for the GAN.  

\subsection{Non-linear regression with Normal errors}
\begin{figure}[H]
\centering
\def\tabularxcolumn#1{m{#1}}

\subfloat[$x=0.1$]{\includegraphics[width = 3cm]{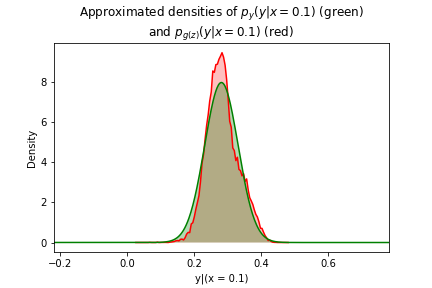}} 
\subfloat[$x=0.4$]{\includegraphics[width = 3cm]{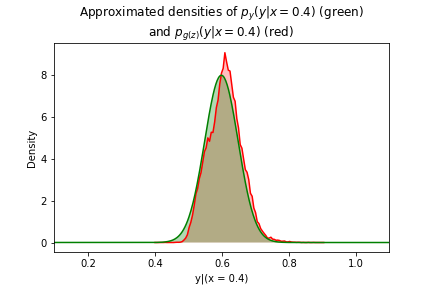}}
\subfloat[$x=0.7$]{\includegraphics[width = 3cm]{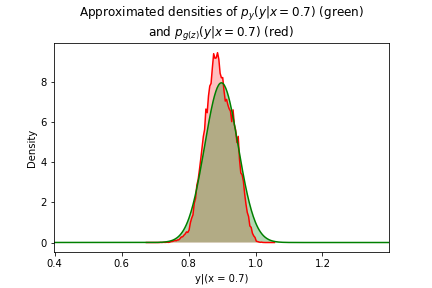}} 
\subfloat[$x=1.0$]{\includegraphics[width = 3cm]{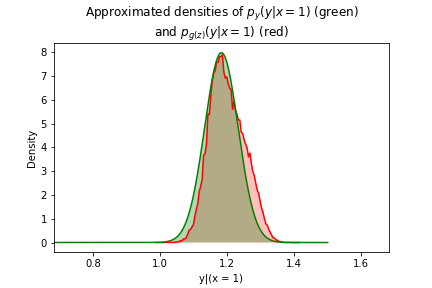}}

\caption{Model 3: True and generated densities}
\label{figure12}
\end{figure}

The third model is a mixture of a sigmoid and a linear trend with Normal errors. It is a significantly more complex function. The comparison between true and generated densities (Figure \ref{figure12}), even if still convincing, show that the generated distribution is not as good as in the previous scenarios. The generated distributions are skewed and even non-centered for $x=1.0$. 

\begin{figure}[H]
\centering
\def\tabularxcolumn#1{m{#1}}

\subfloat[$x=0.1$]{\includegraphics[width = 3cm]{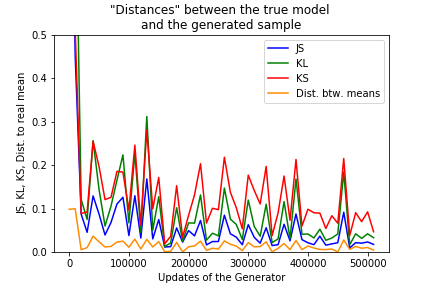}} 
\subfloat[$x=0.4$]{\includegraphics[width = 3cm]{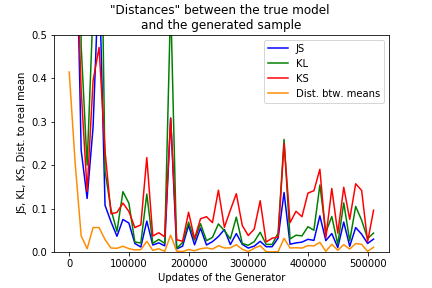}}
\subfloat[$x=0.7$]{\includegraphics[width = 3cm]{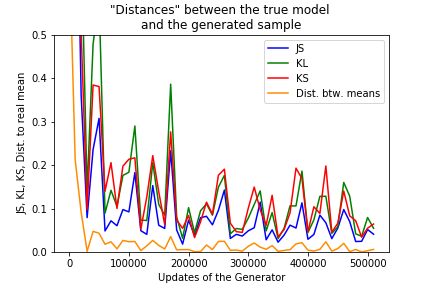}} 
\subfloat[$x=1.0$]{\includegraphics[width = 3cm]{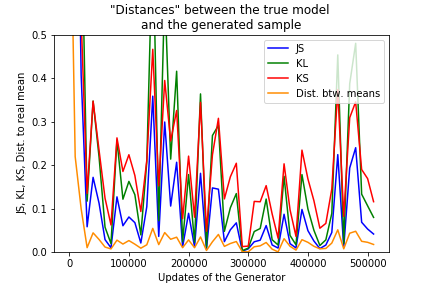}}

\caption{Model 3: Distances between true and generated distributions}
\label{figure13}
\end{figure}

The first impression is confirmed by the analysis of the distances (Figure \ref{figure13}). Again, the distances between distributions are not bad but not as good as previously. We observe that the GAN is also less accurate for $x=1.0$. It makes sense since $1.0$ is on the edge of the support of $x$. The relation being non-linear, we could have expected that this value would be harder to predict.  

\begin{figure}[H]
\centering
\def\tabularxcolumn#1{m{#1}}

\subfloat[Mean]{\includegraphics[width = 3cm]{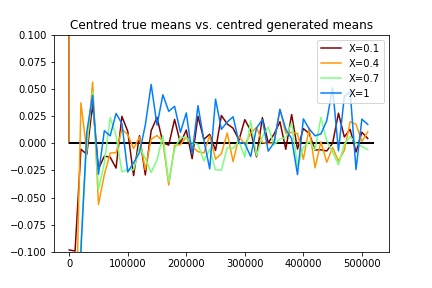}} 
\subfloat[Variance]{\includegraphics[width = 3cm]{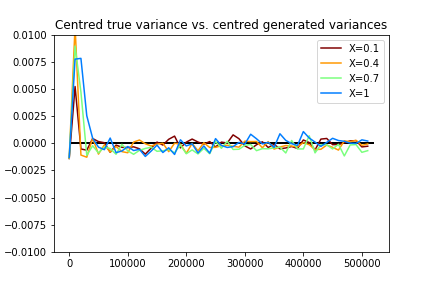}}
\subfloat[Skewness]{\includegraphics[width = 3cm]{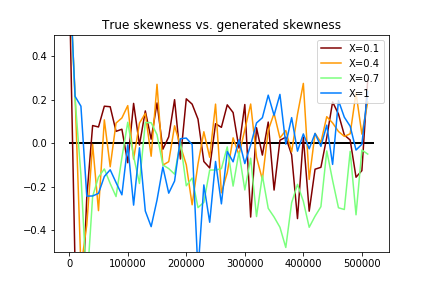}} 
\subfloat[Kurtosis]{\includegraphics[width = 3cm]{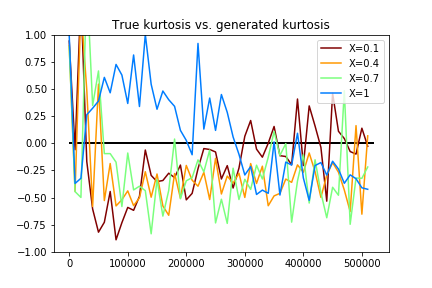}} 
\caption{Model 3: Estimated moments from generated samples compared to true moments}
\label{figure14}
\end{figure}

The difference between the empirical approximate moments and the true moments (Figure \ref{figure14}) corroborates this analysis. The approximations are slightly worse but still converging around the true values. We can also notice that the approximate kurtosis tends to converge slower than it did in the previous scenario. It tells two things. First, it is longer to learn a more complex model. Second, the training of GANs can be long and they can potentially learn still after several hundred thousand of updates.

These first experiments show that GAN can "learn" a conditional distribution and can be used in some cases to approximate the distribution for a new prediction in a bivariate regression. Of course, these examples are very limited. The dimensionality is small, the number of events is important, the predictions are given for values next to values contained in the training set, etc. We will discuss these limitations later. However, it seems that using GAN for regression can make sense. 

\subsection{Reverse non-linear regression}

We would like to end up this first overview of the ability of GANs to approximate distributions for a new prediction with the most complex model. Instead of trying to predict $y$ given $x$, the reverse regression problem (or calibration problem) try to predict $x$ given $y$. In this scenario, we keep then the third model but we try the reversed problem.
It is important to note that this task is much more complex than the previous ones since $y$ is a function with no closed-form and it can have multiple solutions, which leads to potentially multimodal distributions for a given $y$. Since there is no closed-form for it, the true distribution is no longer the analytic one but an approximation by sampling as explained in chapter 3. 

\begin{figure}[H]
\centering
\def\tabularxcolumn#1{m{#1}}

\subfloat[$y=0.1$]{\includegraphics[width = 3cm]{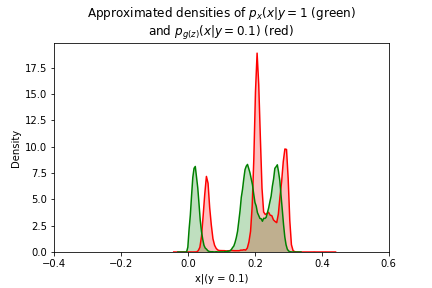}} 
\subfloat[$y=0.4$]{\includegraphics[width = 3cm]{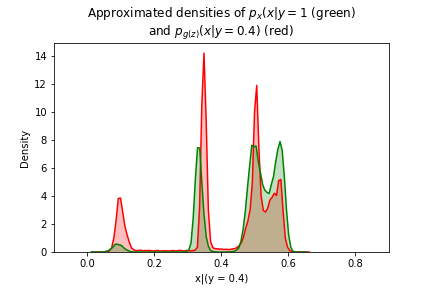}}
\subfloat[$y=0.7$]{\includegraphics[width = 3cm]{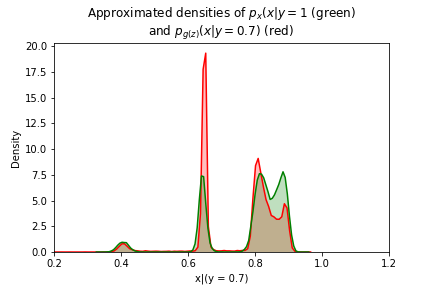}} 
\subfloat[$y=1.0$]{\includegraphics[width = 3cm]{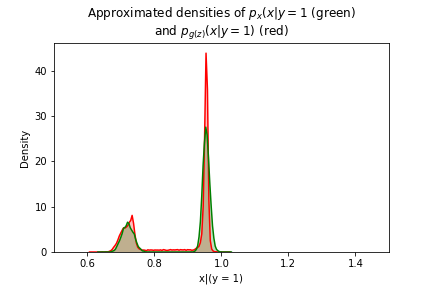}}

\caption{Model 3 - Inverse regression : True and generated densities}
\label{figure15}
\end{figure}

The plots of the true and generated distribution show (figure \ref{figure15}) the complexity of the densities and the difficulty for the GAN to yield good approximations. We observe, in particular, a biased approximation for $y=0.1$, a slightly biased and a partial mode collapse for $x=0.4$, a partial mode collapse for $y=0.7$ and a too concentrated distribution in one mode for $x=1$. However, the generated distributions are around the good values and the approximations are not bad taking into account the complexity of the task. 

\begin{figure}[H]
\centering
\def\tabularxcolumn#1{m{#1}}

\subfloat[$y=0.1$]{\includegraphics[width = 3cm]{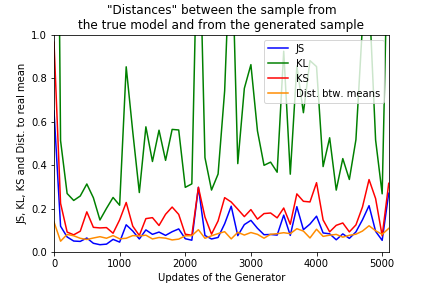}} 
\subfloat[$y=0.4$]{\includegraphics[width = 3cm]{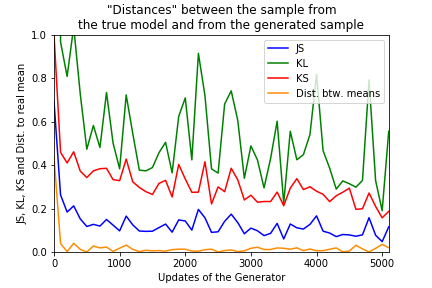}}
\subfloat[$y=0.7$]{\includegraphics[width = 3cm]{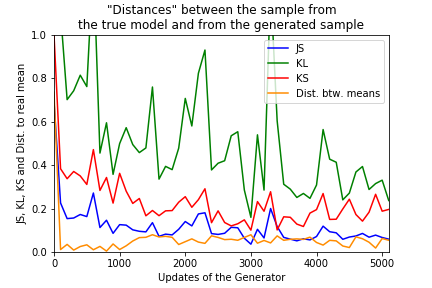}} 
\subfloat[$y=1.0$]{\includegraphics[width = 3cm]{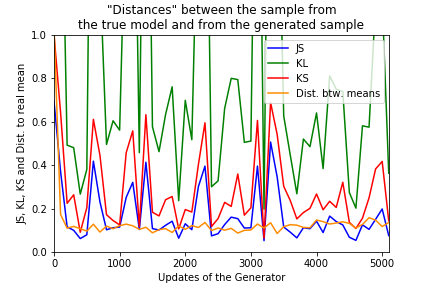}}

\caption{Model 3 - Inverse regression: Distances between true and generated distributions}
\label{figure16}
\end{figure}

 It does a reasonable, though imperfect job, for which taking another training step may help, or adjusting the optimization.
Of course, in comparison with previous experiments, the generated distribution looks far from the real distribution. The distances plotted in Figure \ref{figure16} attest that the approximation is less accurate. 

We let the plots showing the distance between moments at the same scale that we set for the previous regression problem. A lot of values are out of the range. It looks a bit disappointing but again it reflects much more complexity of the task than a lack of performance. 

Of course, this first example of reversed regression is not completely satisfying. However, it supports the idea than adversarial regression can produce interesting results even in complex problems. It encourages to investigates further the quality of GAN. We will analyze more in details the various types of GANs and the effect of the hyperparameters on their performance in adversarial regressions.

\begin{figure}[H]
\centering
\def\tabularxcolumn#1{m{#1}}

\subfloat[Mean]{\includegraphics[width = 3cm]{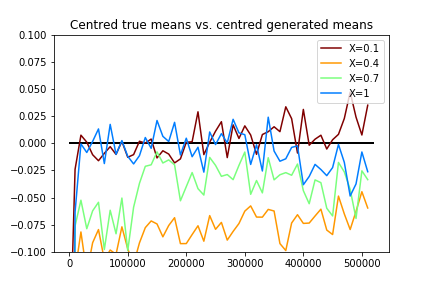}} 
\subfloat[Variance]{\includegraphics[width = 3cm]{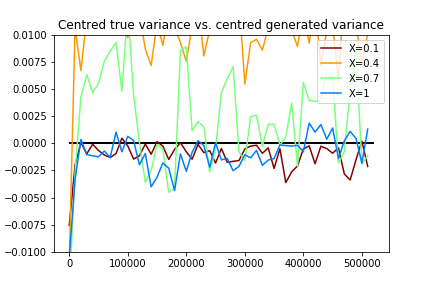}}
\subfloat[Skewness]{\includegraphics[width = 3cm]{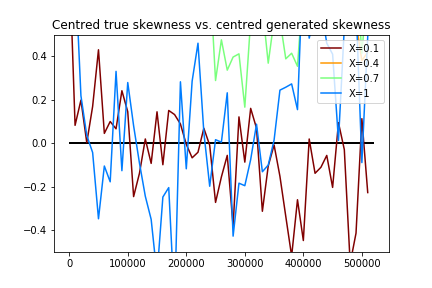}} 
\subfloat[Kurtosis]{\includegraphics[width = 3cm]{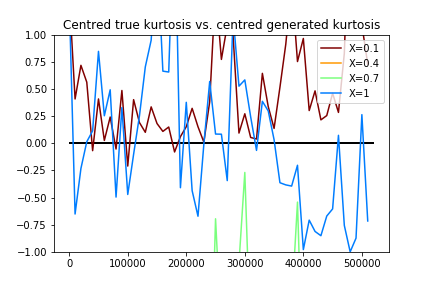}} 
\caption{Model 3 - Inverse regression: Estimated moments from generated samples compared to true moments}
\label{figure17}
\end{figure}

\section{Comparing GANs for adversarial regressions}

Some hyperparameters and experiments conditions, such as the size of the dataset, are directly related to the quality of the approximation. As we will see, the dimension of the noise vector, the size of the sample, the batch size and the type of GAN influence the quality of the regression. We will go over these factors with the idea of finding the good properties of GANs for regression tasks. 

\subsection{The effect of the dimensionality of the noise vector}

As seen in Chapter 2, the Generator requires a random input to generate fake examples. In our case, we choose the standard normal to produce the noise. After having trouble at obtaining a good fit of the true distribution, we have figured out that a one-dimensional noise was the source of the problems. Having a higher dimension for $Z$ was helping a lot. Therefore, we wanted to test the effect of the dimensionality of the noise.

\begin{figure}[H]
\centering
\def\tabularxcolumn#1{m{#1}}
\begin{tabular}{cc}
\subfloat[y=0.1]{\includegraphics[width=5.5cm]{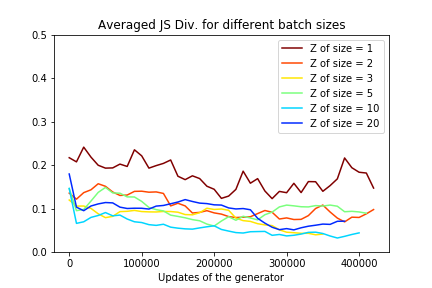}} 
\subfloat[y=0.4]{\includegraphics[width = 5.5cm]{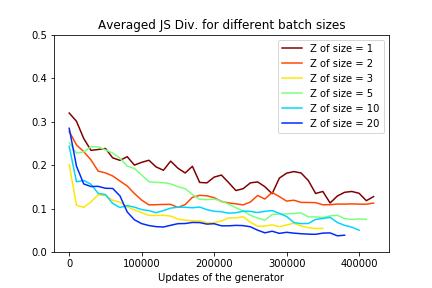}}\\
\subfloat[y=0.7]{\includegraphics[width = 5.5cm]{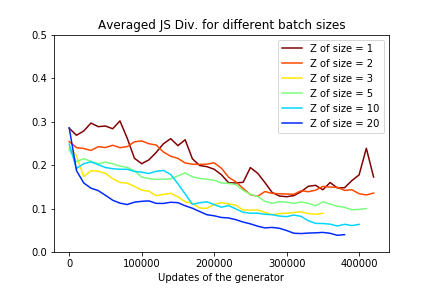}} 
\subfloat[y=1]{\includegraphics[width = 5.5cm]{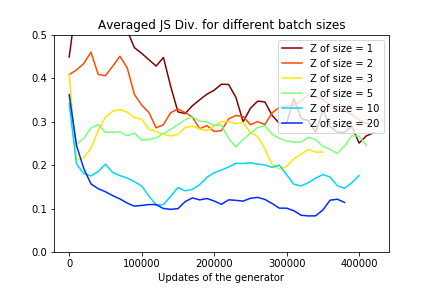}} \end{tabular}
\caption{JS Divergences for various dimensions of the noise}
\label{figureNdim}
\end{figure}

A noise with few dimensions harms the training process. However, it is not clear that a bigger noise is always related with better approximation. We tested noises $Z$ with value 1, 2, 3, 5, 10, and 20. The quality tends to increase with the number of dimension of the noise. It is particularly clear for the lower dimensions of $Z$ (1 and 2). For upper values, even if the trends stays, the improvements are not consistent and if existing low. This trend is illustrated in Figure \ref{figureNdim} representing training for the third experiment of a standard GAN for the various noise dimensions. However, we have found similar results for all the experiments and all the types of GAN. 

\subsection{Adversarial regression and sample size}

A second important factor to consider is the quantity of data required to produce a good approximation of the true distribution. We choose to test the values with samples from the true distribution of size 250, 500, 1000, 2500, 5000, 10000 and 100000. 

\begin{figure}[H]
\centering
\def\tabularxcolumn#1{m{#1}}
\begin{tabular}{cc}
\subfloat[y=0.1]{\includegraphics[width = 5.5cm]{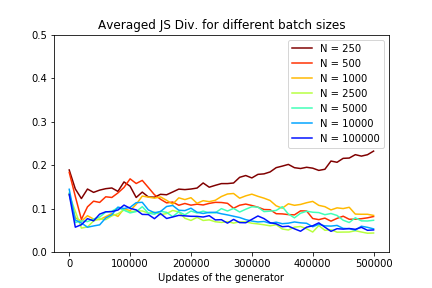}} 
\subfloat[y=0.4]{\includegraphics[width = 5.5cm]{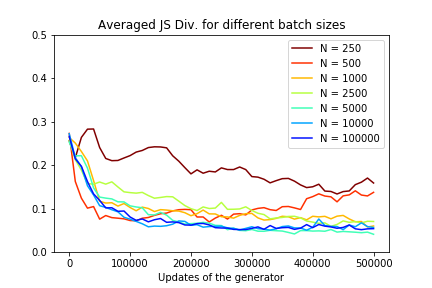}}\\
\subfloat[y=0.7]{\includegraphics[width = 5.5cm]{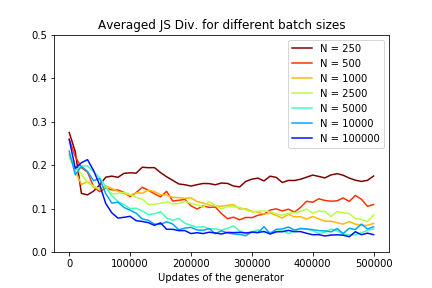}} 
\subfloat[y=1]{\includegraphics[width = 5.5cm]{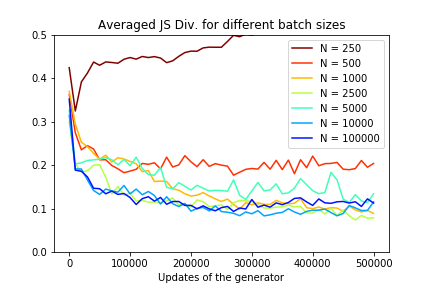}} \end{tabular}
\caption{JS Divergences for various sample sizes}
\label{figure32}
\end{figure}

We have found similar results to the ones about the dimensionality of the noise. It appears that a small sample prevents the models to make good approximations of the true distribution. However, as the size of the sample increases, the improvements are reduced and eventually stagnated. The figure \ref{figure32}, representing the training of a standard GANs for the third experiment with different data sizes, illustrates this. The sample of 250 data produces poor approximation, while after 1000 examples the improvements are very limited. Again, we have tested more types of GAN for all three experiments with similar results.

\subsection{comparing batch sizes for different GAN}

In this part, we want to find, if not the best at least a good batch size for each type of GANs. The aim is to find for each GAN, which batch size performs the best. So that we could then compare the best GANs between them.

\subsection{The first two experiments}

As we are going to see, the quality of the generated distributions makes hard to extrapolate general conclusions from these two first experiments. It is, however, possible to identify some trends. 

First observation, the small batch sizes fail at finding a good solution. For all the types of GAN, all the experiments and all the tested points the batch of 20 events are the further away from the true distribution and so during the main part of the training after few updates.
The batch of 80 is not as consistently bad but it is not performing as good as the other. For the bigger batches, it seems than the bigger produces better results but the trend is not consistent and the distances are close to each other.  

\begin{figure}[H]
\centering
\def\tabularxcolumn#1{m{#1}}
\subfloat[$y=0.1$]{\includegraphics[width = 3cm]{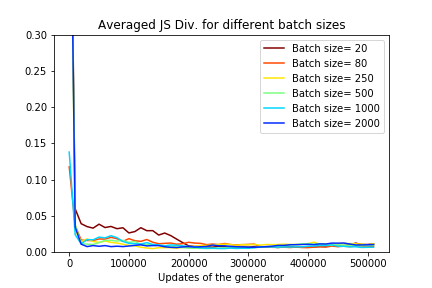}} 
\subfloat[$y=0.4$]{\includegraphics[width = 3cm]{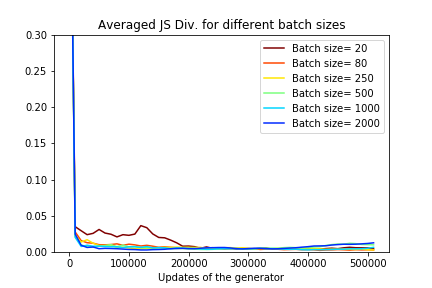}}
\subfloat[$y=0.7$]{\includegraphics[width = 3cm]{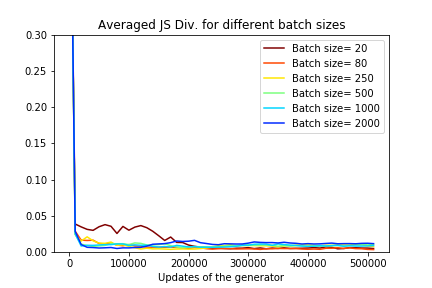}}
\subfloat[$y=1.0$]{\includegraphics[width = 3cm]{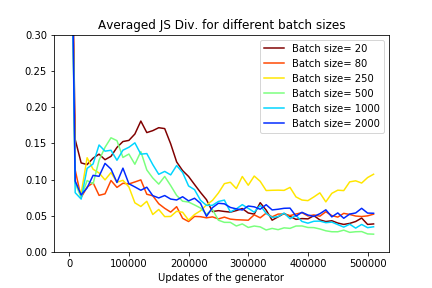}}
\caption{SGAN - Model 1: JS divergence between true and generated distributions}
\label{SGAN1}
\end{figure}

\begin{figure}[H]
\centering
\def\tabularxcolumn#1{m{#1}}
\subfloat[$y=0.1$]{\includegraphics[width = 3cm]{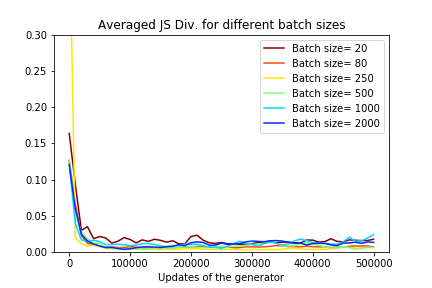}} 
\subfloat[$y=0.4$]{\includegraphics[width = 3cm]{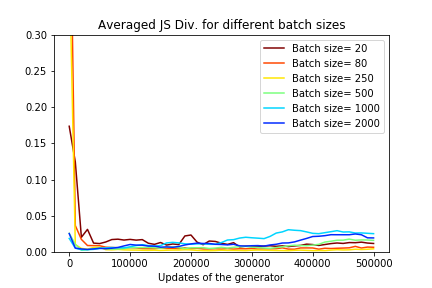}}
\subfloat[$y=0.7$]{\includegraphics[width = 3cm]{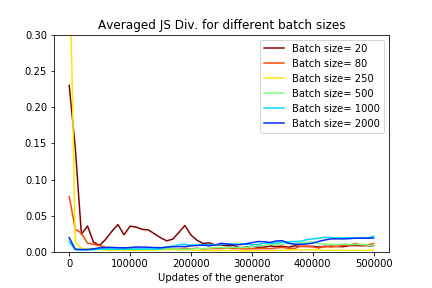}} 
\subfloat[$y=1.0$]{\includegraphics[width = 3cm]{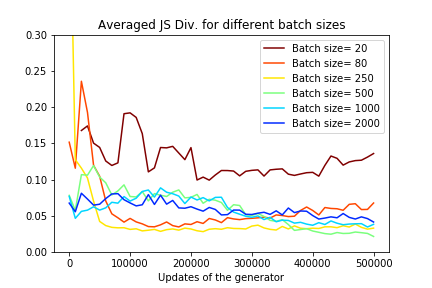}}
\caption{SGAN - Model 2:JS divergence between true and generated distributions}
\label{SGAN2}
\end{figure}

Figures \ref{SGAN1} and \ref{SGAN2} show the training of the SGAN for various batch sizes. These two first experiments give few pieces of information about the impact of the batch size on the quality of the adversarial regression. We observe that, in general, the distance with the true distribution is very small. The batch of 20 events needs slightly more updates to reach values comparable to the other batch sizes. It can be explained by the fact that it is learning with much less information. Training with a batch size of 20 goes through the whole sample every 500 updates, while a batch size of 2000 does it in only 5 updates.

\begin{figure}[H]
\centering
\def\tabularxcolumn#1{m{#1}}
\subfloat[$y=0.1$]{\includegraphics[width = 3cm]{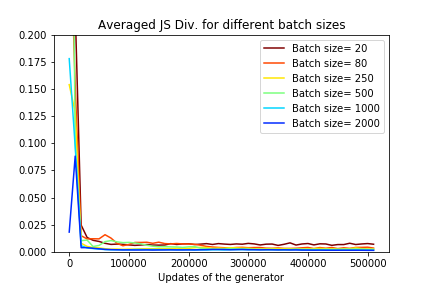}} 
\subfloat[$y=0.4$]{\includegraphics[width = 3cm]{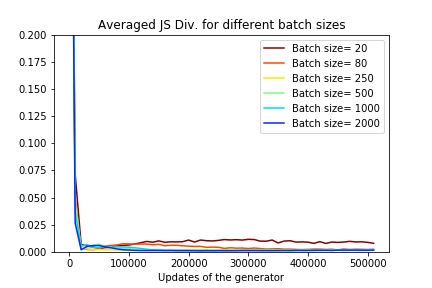}}
\subfloat[$y=0.7$]{\includegraphics[width = 3cm]{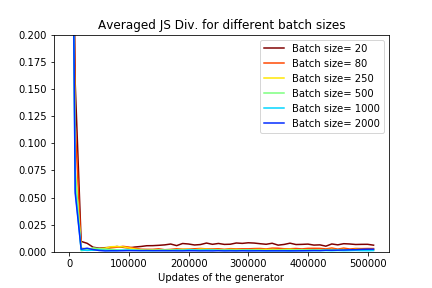}} 
\subfloat[$y=1.0$]{\includegraphics[width = 3cm]{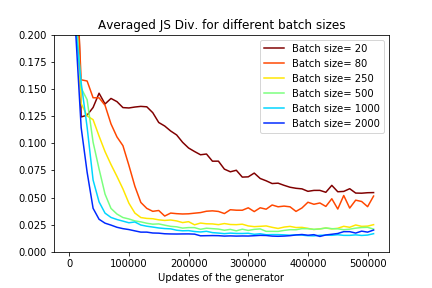}}
\caption{WGAN - Model 1: JS divergence between true and generated distributions}
\label{WGAN1}
\end{figure}

\begin{figure}[H]
\centering
\def\tabularxcolumn#1{m{#1}}
\subfloat[$y=0.1$]{\includegraphics[width = 3cm]{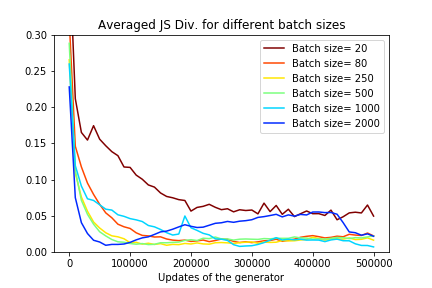}}
\subfloat[$y=0.4$]{\includegraphics[width = 3cm]{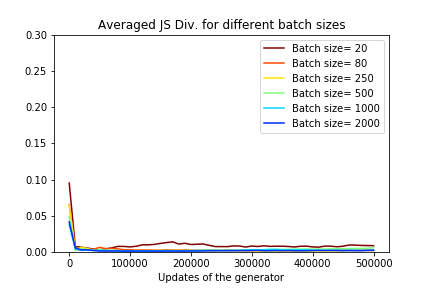}}
\subfloat[$y=0.7$]{\includegraphics[width = 3cm]{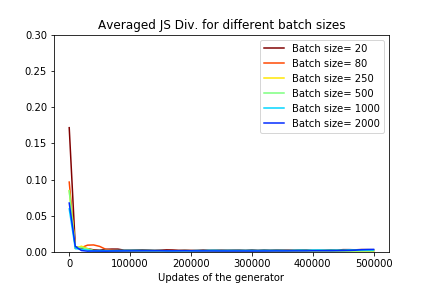}}
\subfloat[$y=1.0$]{\includegraphics[width = 3cm]{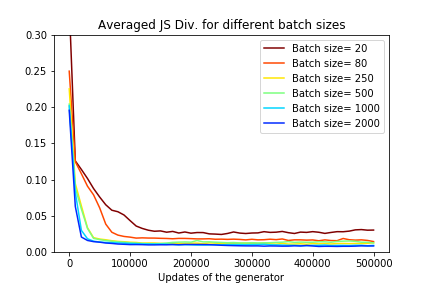}}
\caption{WGAN - Model 2: JS divergence between true and generated distributions}
\label{WGAN2}
\end{figure}

The same experiment with the Wasserstein GAN gives similar results. We can observe it in Figures \ref{WGAN1} and \ref{WGAN2}.  Also, we can notice a small trend: It seems that the bigger batch sizes perform better than the smaller ones. It is not very consistent results and the distances are sometimes too close to make the differences significant, but still, a tendency seems to appear.

\begin{figure}[H]
\centering
\def\tabularxcolumn#1{m{#1}}
\begin{tabular}{cc}
\subfloat[$x=0.1$]{\includegraphics[width = 3cm]{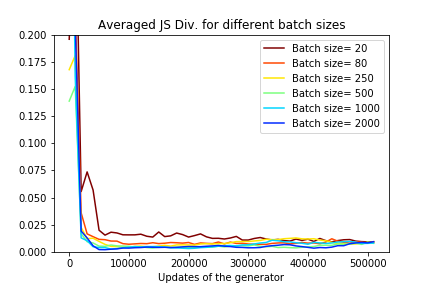}}
\subfloat[$x=0.4$]{\includegraphics[width = 3cm]{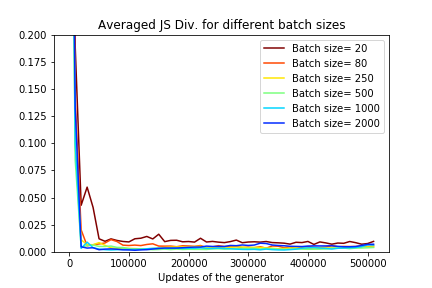}}
\subfloat[$x=0.7$]{\includegraphics[width = 3cm]{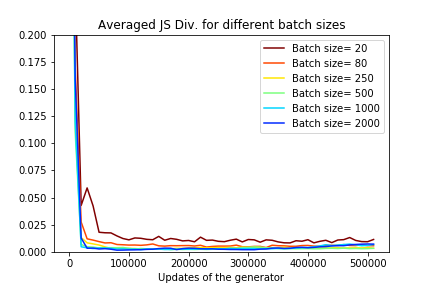}} 
\subfloat[$x=1.0$]{\includegraphics[width = 3cm]{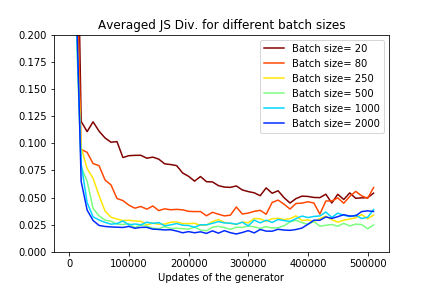}}

\end{tabular}
\caption{RSGAN - Model 1: JS divergence between true and generated distributions}
\label{RSGAN1}
\end{figure}

\begin{figure}[H]
\centering
\def\tabularxcolumn#1{m{#1}}
\subfloat[$x=0.1$]{\includegraphics[width = 3cm]{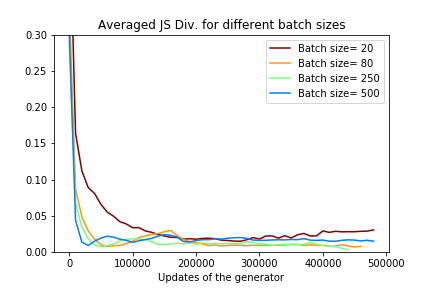}} 
\subfloat[$x=0.4$]{\includegraphics[width = 3cm]{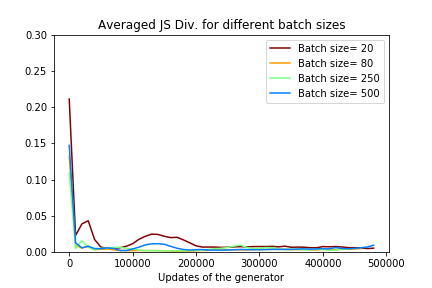}}
\subfloat[$x=0.7$]{\includegraphics[width = 3cm]{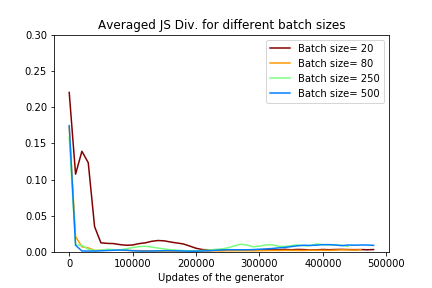}} 
\subfloat[$x=1.0$]{\includegraphics[width = 3cm]{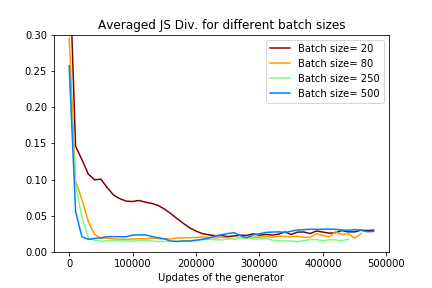}}
\caption{RSGAN - Model 2: JS divergence between true and generated distributions}
\label{RSGAN2}
\end{figure}

We observe again similar results for the Relativistic GAN. The small batch sizes require more updates. They tend also of being a bit further from the true distribution. The plots in Figures \ref{RSGAN1} and \ref{RSGAN2} reports these results. However, the approximation is generally good. Even the worse examples produce distributions close to the real one. The good quality of all the example make harder to analyze the differences. 

From these two first experiments, we can draw some conclusions by comparing the results for the three GAN. First, all the tested GANs perform well for these experiments. The distances with the true distribution are very low. We can also notice that all the three GANs make worse predictions when $y=1$. As explained above, it comes from the fact that the value is on the edge of the distribution and that the distribution is not continuous.
We also can observe than the WGAN and the RSGAN are not performing as well for $y=0.1$ than for $y=0.4$ and $y=0.7$ in the second experiment. This can result from the very concentrated distribution at that point. It means that a small bias for that value can result in a large effect on the distance. 
Finally, the differences appear much clearer when the training is harder. It is the reason why the third experiment should give more information. 

\subsection{The third experiments}

As we have seen, the third experiment is much more complex and involves multimodal distributions. This complexity makes appear the differences with more clarity because the results are more contrasted. Therefore, we will look more carefully at this experiment than at the previous one.

\begin{figure}[H]
\centering
\def\tabularxcolumn#1{m{#1}}
\begin{tabular}{cc}
\subfloat[y = 0.1]{\includegraphics[width =5.5cm]{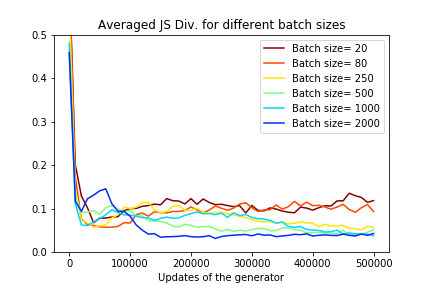}} 
\subfloat[y = 0.4]{\includegraphics[width =5.5cm]{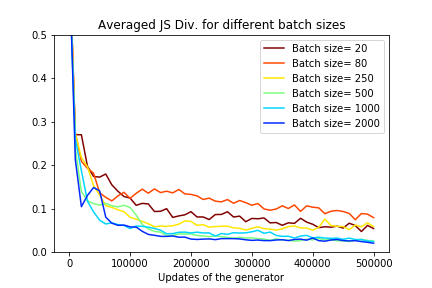}} \\
\subfloat[y = 0.7]{\includegraphics[width =5.5cm]{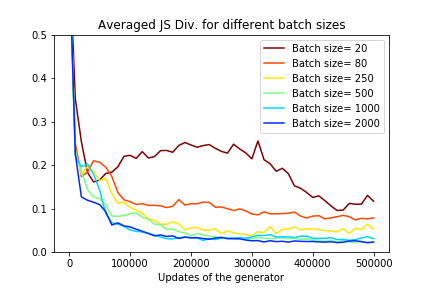}} 
\subfloat[y = 1]{\includegraphics[width =5.5 cm]{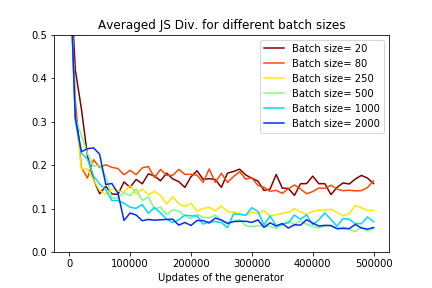}} 
\end{tabular}
\caption{SGAN - Model 3: JS divergence between true and generated distributions}
\label{figure18}
\end{figure}

Let start with the standard GAN. Figure \ref{figure18} presents the evolution of the JS divergence for the 4 $y$'s considered. In this case, the effect of the batch size appears clearly. The bigger the batch size, the better the generated sample. 

\subsubsection{WGAN-GP}

The Wasserstein GAN with gradient penalization presents similar features. Again, the bigger the batch size, the lower the distance. We also observe than the differences between batch sizes are bigger early in the training and tend to diminish later. This trend is much more consistent than for the two first experiment. The smaller batch is very often the worse while the bigger batch is almost always the best.

\begin{figure}[H]
\centering
\def\tabularxcolumn#1{m{#1}}
\begin{tabular}{cc}
\subfloat[$x=0.1$]{\includegraphics[width = 5.5cm]{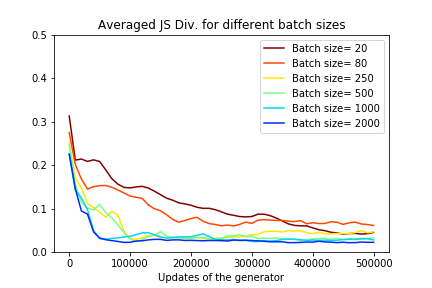}} &
\subfloat[$x=0.4$]{\includegraphics[width = 5.5cm]{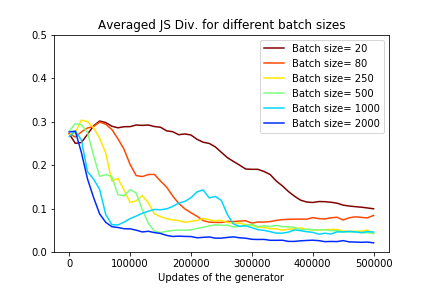}}\\
\subfloat[$x=0.7$]{\includegraphics[width = 5.5cm]{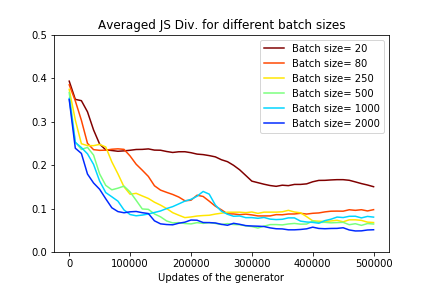}} &
\subfloat[$x=1.0$]{\includegraphics[width = 5.5cm]{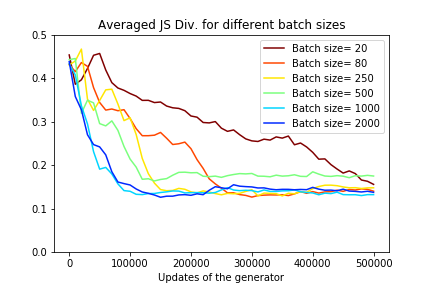}}

\end{tabular}
\caption{WGAN - Model 3: JS divergence between true and generated distributions}
\label{WGAN model3}
\end{figure}

\subsubsection{Relativistic GAN}
The relativistic GAN is the last model tasted. First of all, the training with the batch of size 20 did not converge. So, it will not appears in the plots. It can mean that the convergence is a bit less stable for that GAN. 
As in the other models, in the first part of the training, the larger batch sizes seems better. However, in the second half of the training, all the distances converge to very similar values. Except for the batch of 250 events for the last observation they all converge to very close values. Because we have to pick one for the comparison will take the RSGAN with a batch of 500 which has the lowest sum of distances.

\begin{figure}[H]
\centering
\def\tabularxcolumn#1{m{#1}}
\begin{tabular}{cc}
\subfloat[$y=0.1$]{\includegraphics[width = 5.5cm]{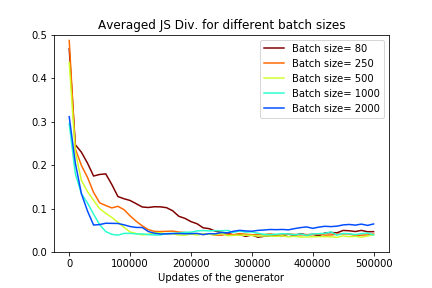}} &
\subfloat[$y=0.4$]{\includegraphics[width = 5.5cm]{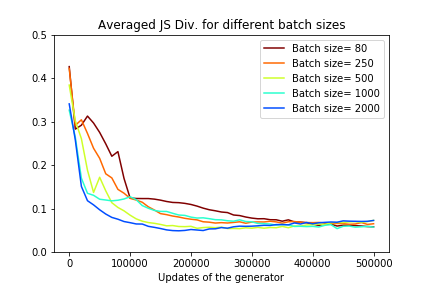}}\\
\subfloat[$y=0.7$]{\includegraphics[width = 5.5cm]{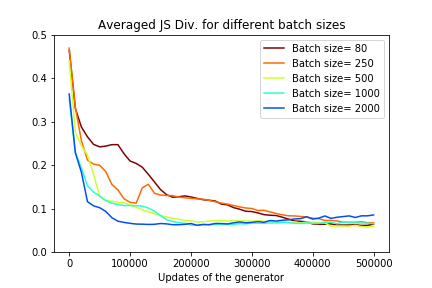}} &
\subfloat[$y=1.0$]{\includegraphics[width = 5.5cm]{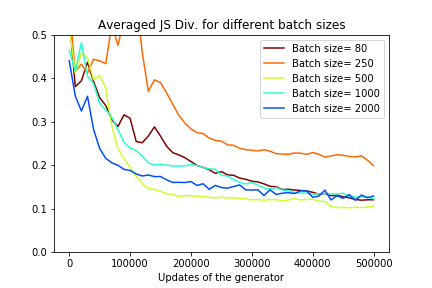}}

\end{tabular}
\caption{RSGAN - Model 3: JS divergence between true and generated distributions}
\label{RSGANJSDIV}
\end{figure}

\subsection{Are GANs equal?}

\begin{figure}[H]
\centering
\def\tabularxcolumn#1{m{#1}}
\begin{tabular}{cc}
\subfloat[y = 0.1]{\includegraphics[width =5.5cm]{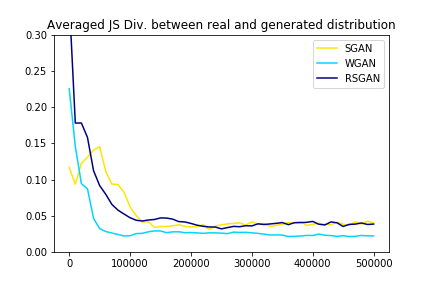}} &
\subfloat[y = 0.4]{\includegraphics[width = 5.5cm]{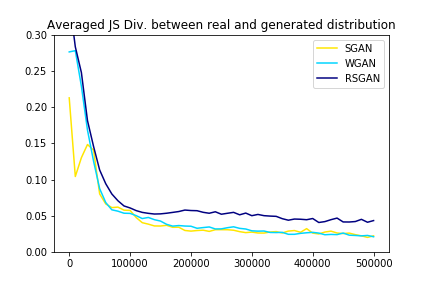}} \\
\subfloat[y = 0.7]{\includegraphics[width = 5.5cm]{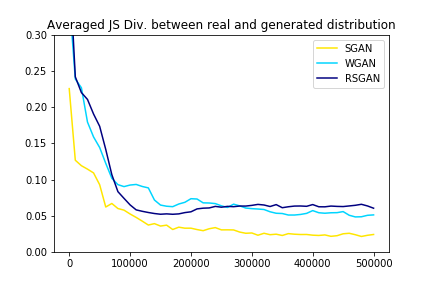}} &
\subfloat[y = 1]{\includegraphics[width = 5.5cm]{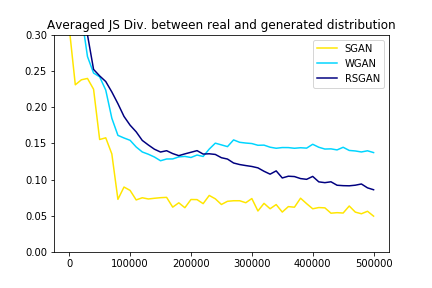}} 
\end{tabular}
\caption{Comparison between GANs: Averaged JS divergences}
\label{figure28}
\end{figure}

We have found good batch size values for the three GAN tested: 2000 for the SGAN, 2000 for the WGAN-GP and 500 for the RSGAN. We can, therefore, compare these models. Figure \ref{figure28} compares the distances for the three GANs. The first information is that they all produce approximations in the same order of magnitude. None of them shows performances highly superior or inferior to the other. No GAN is consistently better for all the observation either.
 
The comparison between the averaged distance reveals slightly better performances for the Standard GAN. The case of $y=0.1$ is the only measure where the WGAN perform slightly better. They perform both as well for $y=0.4$. For the two last values, the Standard GAN performs much better than the others at every step of the training.

In the opposite, the RSGAN produces worse samples than the two other models expect at the end of the training for $y=0.1$. This result is surprising since the RSGAN is supposed to be an improvement of the Standard GAN. However, it doesn't mean that there is an error neither in most of the papers nor in this study. The result can come from the very specific condition of experimentation of this study. In particular, it can be that RSGAN performs comparatively better in high dimensions.

\begin{figure}[H]
\centering
\def\tabularxcolumn#1{m{#1}}
\begin{tabular}{cc}
\subfloat[y = 0.1]{\includegraphics[width =5.5cm]{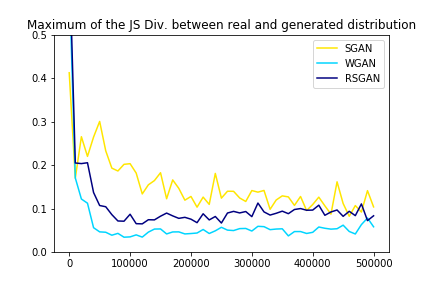}} &
\subfloat[y = 0.4]{\includegraphics[width = 5.5cm]{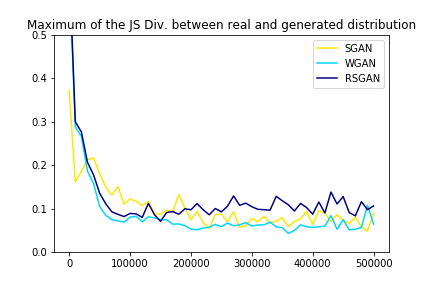}}\\
\subfloat[y = 0.7]{\includegraphics[width = 5.5cm]{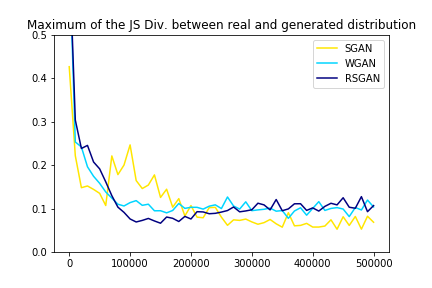}} &
\subfloat[y = 1]{\includegraphics[width = 5.5cm]{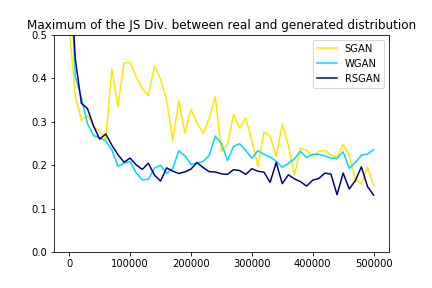}} \end{tabular}
\caption{Comparison between GANs: Maximum of the JS divergences}
\label{figure29}
\end{figure}

Since the variances of the distance are not similar between the different types of GAN, it is interesting to compare the maximum distance between them. The plots of the maximum distance plotted in Figure \ref{figure29} shows results even closer and ambiguous. With this metric, no GAN looks better than another. The reason is that the SGAN varies in broader range during the training. Because of this higher variance, it loses the small advantages it had with the averaged divergence. 

Let summarize the main founding on the part by putting together the advantages and drawbacks of the three GANs. 

The SGAN has better allover performance. In this kind of task, it seems to reach a lower minimum on average. It has also the quality of being easier conceptually. The WGAN-GP has performance slightly less good. However, its lower variance makes of it a safe choice. Though, we would have except better results from this type, since it is presented as the state-of-the-art loss function. Finally, the Relativistic GAN performs well but slightly worse than the two other GANs. However, it is the fastest to train since it does not require several Discriminator updates before an update of the Generator. Moreover, RSGAN is stable during training. The influence of the batch size on the performance is very low. Almost all the batch sizes converged around the same value. Also, the variance during the training is low, so much so that it presents similar results than the other GAN for the maximal distance. The main drawback is the average performance. In average, the RSGAN is systematically worse than the SGAN. All these properties make of the RSGAN a safe choice, not producing sample surprisingly far from the real distribution. However, on average, there is a better solution.
Surprisingly, the best GAN found is the original one. Some of the most famous and promising improvement of the SGAN are not able to perform better in these experiments. The variability of the training is its only drawback.

\begin{figure}[H]
\centering
\subfloat[y = 0.1]{\includegraphics[width =3cm]{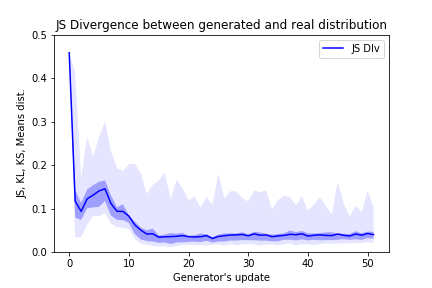}} 
\subfloat[y = 0.4]{\includegraphics[width = 3cm]{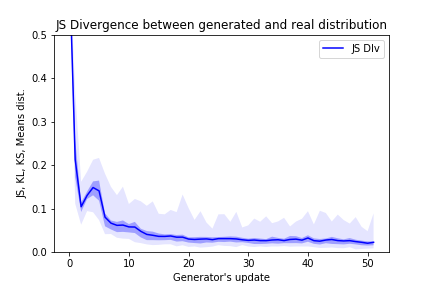}}
\subfloat[y = 0.7]{\includegraphics[width = 3cm]{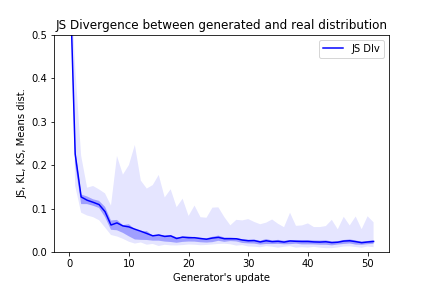}}
\subfloat[y = 1]{\includegraphics[width = 3cm]{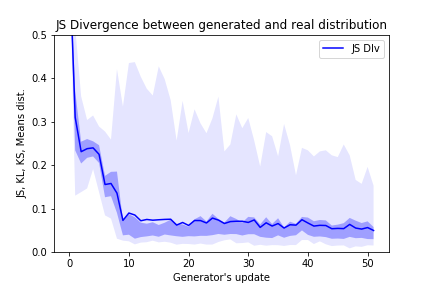}} 
\caption{JS Divergences for the best model}
\label{Best_dist}
\end{figure}

\begin{figure}[H]
\centering
\subfloat[y = 0.1]{\includegraphics[width =3cm]{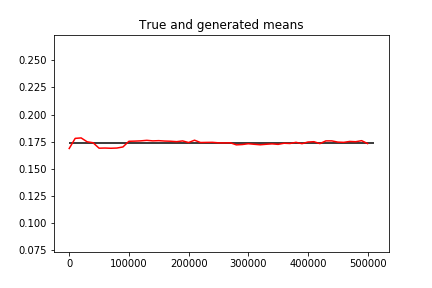}} 
\subfloat[y = 0.4]{\includegraphics[width = 3cm]{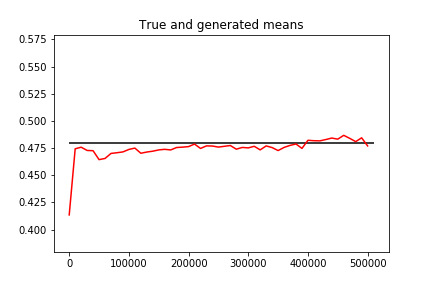}}
\subfloat[y = 0.7]{\includegraphics[width = 3cm]{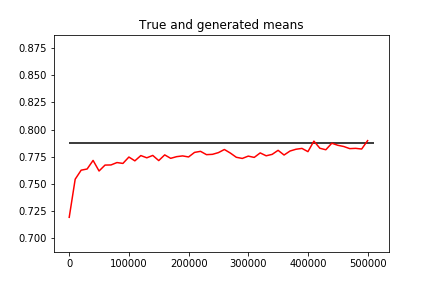}}
\subfloat[y = 1]{\includegraphics[width = 3cm]{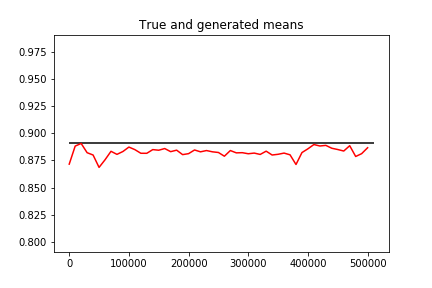}} 
\caption{Distance between means for the best model}
\label{Best_means}
\end{figure}

\begin{figure}[H]
\centering
\subfloat[y = 0.1]{\includegraphics[width =3cm]{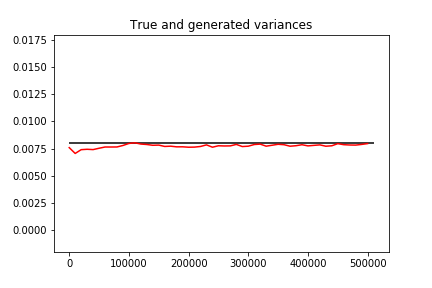}} 
\subfloat[y = 0.4]{\includegraphics[width = 3cm]{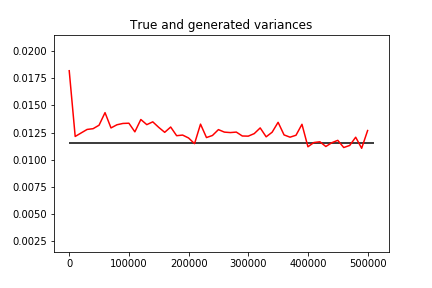}}
\subfloat[y = 0.7]{\includegraphics[width = 3cm]{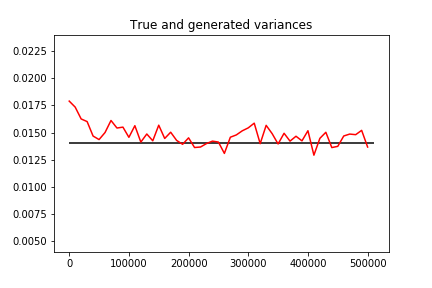}}
\subfloat[y = 1]{\includegraphics[width = 3cm]{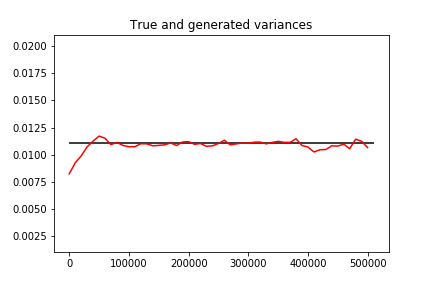}} 
\caption{Distance between variances for the best model}
\label{Best_var}
\end{figure}

\begin{figure}[H]
\centering
\subfloat[y = 0.1]{\includegraphics[width =3cm]{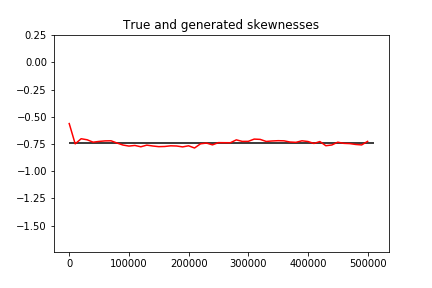}} 
\subfloat[y = 0.4]{\includegraphics[width = 3cm]{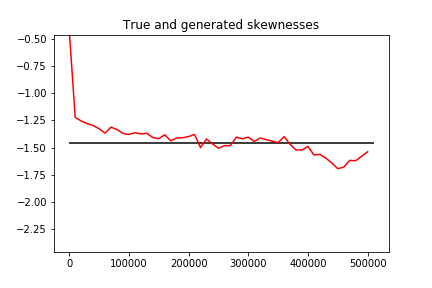}}
\subfloat[y = 0.7]{\includegraphics[width = 3cm]{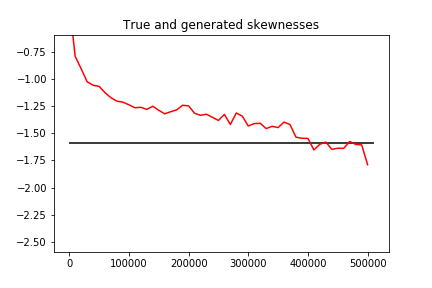}}
\subfloat[y = 1]{\includegraphics[width = 3cm]{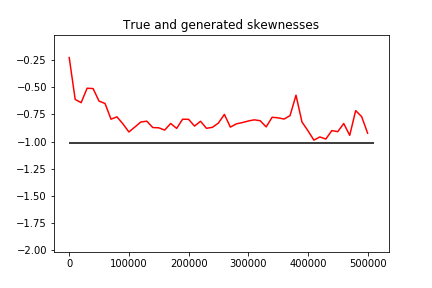}} 
\caption{Distance between skewnesses for the best model}
\label{Best_skew}
\end{figure}

\begin{figure}[H]
\centering
\subfloat[y = 0.1]{\includegraphics[width =3cm]{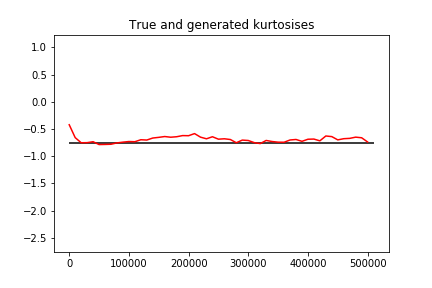}} 
\subfloat[y = 0.4]{\includegraphics[width = 3cm]{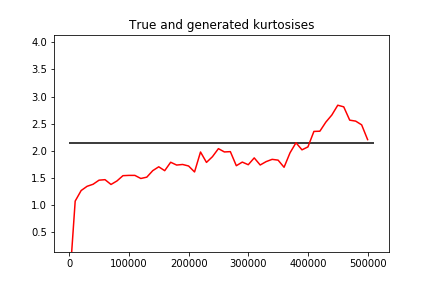}}
\subfloat[y = 0.7]{\includegraphics[width = 3cm]{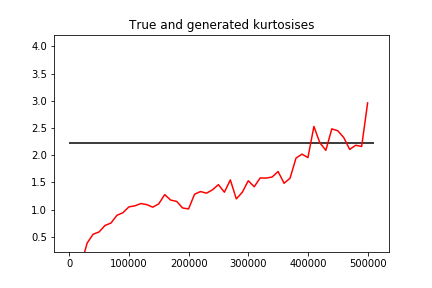}}
\subfloat[y = 1]{\includegraphics[width = 3cm]{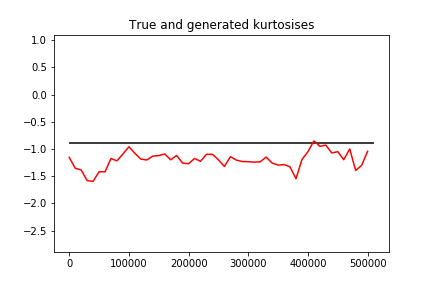}} 
\caption{Distance between kurtosises for the best model}
\label{Best_kurt}
\end{figure}

\begin{figure}[H]
\centering
\subfloat[y = 0.1]{\includegraphics[width =3cm]{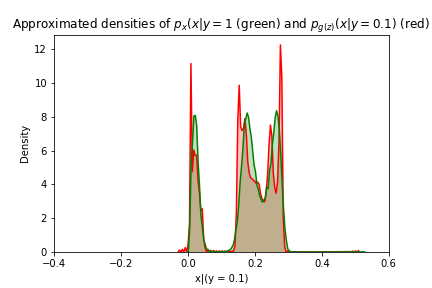}} 
\subfloat[y = 0.4]{\includegraphics[width = 3cm]{Figures/Best/Best_Density_0.png}}
\subfloat[y = 0.7]{\includegraphics[width = 3cm]{Figures/Best/Best_Density_0.png}}
\subfloat[y = 1]{\includegraphics[width = 3cm]{Figures/Best/Best_Density_0.png}} 
\caption{Compared densities for the best model}
\label{Best_Dens}
\end{figure}

The previous sections of this chapter helped us to find the best models of our experiments. In particular, we ended up with the SGAN performing slightly better than the others. The comparison between GANs required metrics that can be quite abstract and their meaning about the real closeness between distributions hard to visualize. The JS Divergence is not easy to interpret intuitively. It is therefore interesting of having a look at some of the distributions produce by the best model and the quality of the statistics generated by its samples so that we have a better idea of what the distance represent. Figure \ref{Best_dist} shows the evolution of the JS Divergence. The range between the minimum and maximum values measured is in light blue and slightly darker is the interquartile range. The line is the average JS DIvergence. Figures \ref{Best_means} to \ref{Best_kurt} compares moments from the true and the generated distributions. Eventually, Figure \ref{Best_Dens} plots true and generated densities. 

The level of approximation reach by this model is remarkable. Even for this complexe multimodal model, it is able to approximate the forth moment. The comparison between densities are remarkable. Even if not perfect, the generated densities find all the modes and submodes. The result is impressive and encouraging. 

\section{Some results of exploratory research}

After all those systematic results, we would like to close this chapter with two results from exploratory research. Even if they suffer of a lack of systematicity, they will be useful for the discussion and the conclusion. The first one concerns the improvement of the adversarial regression. The second concerns the extension of the adversarial regression to higher dimensions. 

As we have seen, the training of GAN is somehow unstable. Updates after updates, the generated sample is always close but never match perfectly the true distribution. In particular, the approximated distribution is almost always more concentrated than the real one. Empirically, it is then tempting to let fluctuate the GAN around the true value and sum up the samples to obtain a better fit. 

\begin{figure}[H]
\centering
\def\tabularxcolumn#1{m{#1}}
\begin{tabular}{cc}
\subfloat[y = 0.1]{\includegraphics[width =5.5cm]{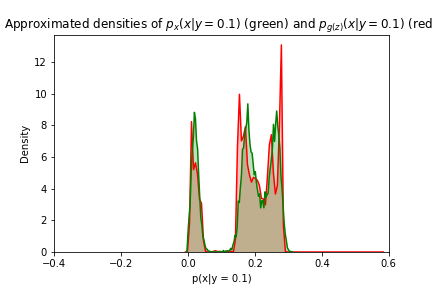}} &
\subfloat[y = 0.1]{\includegraphics[width = 5.5cm]{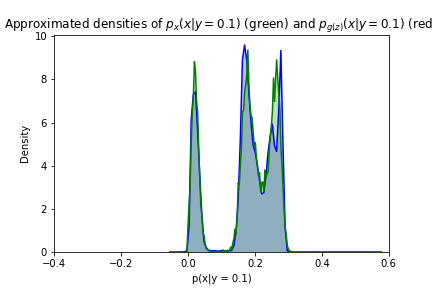}}\\
\subfloat[y = 0.1]{\includegraphics[width =5.5cm]{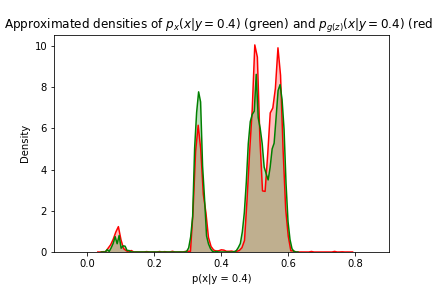}} &
\subfloat[y = 0.1]{\includegraphics[width = 5.5cm]{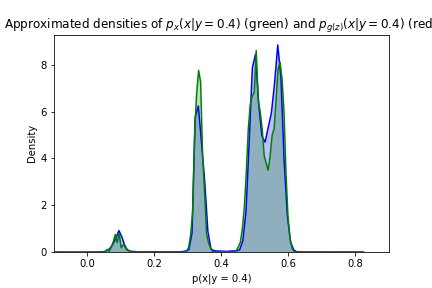}}\\
\subfloat[y = 0.1]{\includegraphics[width =5.5cm]{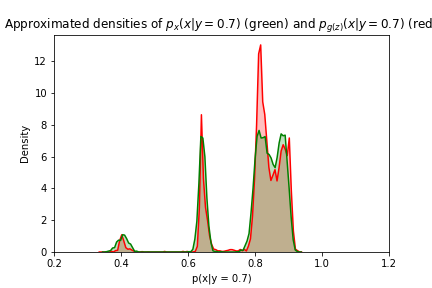}} &
\subfloat[y = 0.1]{\includegraphics[width = 5.5cm]{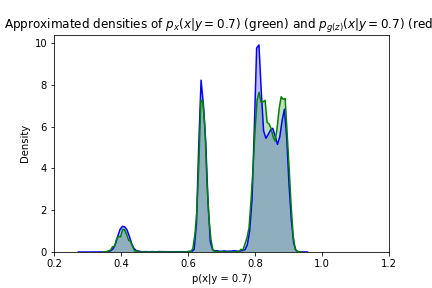}}\\
\subfloat[y = 0.1]{\includegraphics[width =5.5cm]{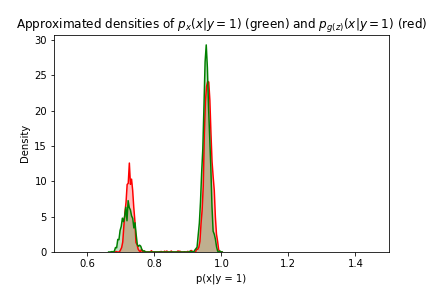}} &
\subfloat[y = 0.1]{\includegraphics[width = 5.5cm]{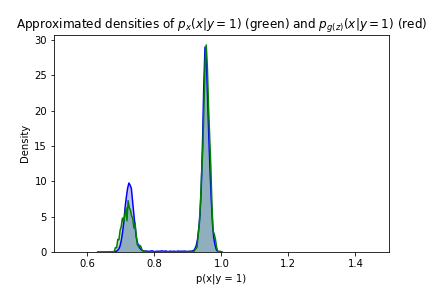}}\\
\end{tabular}
\caption{Comparison between the two approximations of the true distribution}
\label{figure30}
\end{figure}

In Figure \ref{figure30}, we plot the distribution generated from the best results that we obtained (the SGAN with a batch size of 2000). On the left, the red curve represents the generated sample after the 510'000 updates. On the right, the blue curve represents a sum of samples from 20 different states of the Generator at different updates during training (from 310'000 to 510'000). Empirically, this approximation with cumulative samples from the Generator at different steps of the training brings most of the time an important improvement of the adversarial regression. Of course, some theoretical backgrounds and more empirical experiences are needed to support this idea. Other techniques could be developed. This example shows however that GANs are new techniques and they have still a good margin of progression. 

The second result from exploratory research concerns the adversarial regression in higher dimensions. In the presentation of GAN in the first chapter, we have seen that they can potentially produce an approximation of the distribution for a new prediction and non-linear regression in high dimensional space. We have tested the non-linearity but not the high dimensional space. Therefore, we would like to produce few plots about an experiment with $x$ being a vector of size 5.

We defined a non-linear model as follow: 

\[y= \frac{ x_1 + x_2^2 + ln(x_3+1) + 0.2 \cdot sin(20 \cdot x_4) - X_5 + 1 }{3.5} + \epsilon \] 

where, $\epsilon \sim \mathcal{N}(0, 0.05^2)$. 

The model is non-linear but the $y$ is easily predictable. Since the error term follows a one-dimensional normal distribution, the distribution of the prediction for a new $x$ is easy to compute. We plot the true distribution and the generated distribution for four values of $X$. For simplicity, we have kept the same values:  $0.1 \cdot \mathbf{1}_5,0.4 \cdot \mathbf{1}_5, 0.7 \cdot \mathbf{1}_5, \mathbf{1}_5$.

\begin{figure}[H]
\centering
\def\tabularxcolumn#1{m{#1}}
\begin{tabular}{cc}
\subfloat[x = 0.1$_5$]{\includegraphics[width =5.5cm]{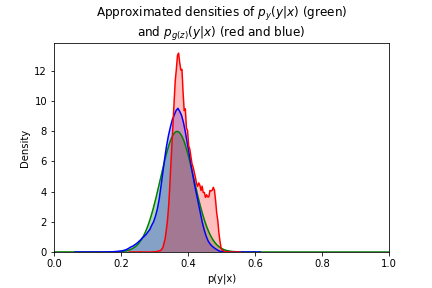}} &
\subfloat[x = 0.4$_5$]{\includegraphics[width = 5.5cm]{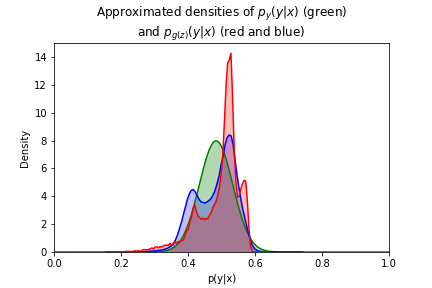}}\\
\subfloat[x = 0.7$_5$]{\includegraphics[width = 5.5cm]{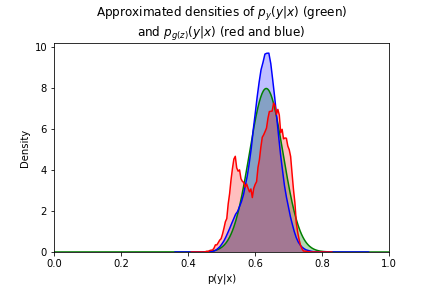}} &
\subfloat[x = 1$_5$]{\includegraphics[width = 5.5cm]{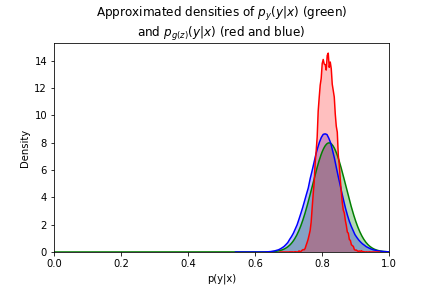}} \end{tabular}
\caption{Generated and true densities with a sample size of 100'000}
\label{figure45}
\end{figure}

Figure \ref{figure45} shows the results for a SGAN with 100'000 events in the dataset: The red distribution is produced by the Generator after 510'000 updates.  The blue is the distribution with 20 samples from the different updates of the Generator as explained previously.

Again, this is only an exploratory experiment. However, it shows that adversarial regression can produce an approximation for a new observation in higher-dimensional spaces. The quality of this approximation requires further study. 

These two examples show that adversarial regression have still an important potential of improvement and second that, in some conditions, they also work in higher dimensions. In the next section, we will discuss the findings of this study enlightened by those two cases.

\section{Promising results}

We havestudied a bunch of parameters influencing the training and the level of convergence of GANs. We looked at various experiments and among them at various points of observation. We changed the dimension of the noise and the sample size. Eventually, we tested various loss functions giving different names to the GANs. 

The first main conclusion of all these experiments is that there is no feature producing a GAN consistently better than the other. In all these tests the differences appear as trends rather than definitive results. This observation says two things. First, the training of GANs is unstable. The variations during training make that no GAN outperforms consistently the others. Second, the quality of GAN is about fine-tuning a range of hyperparameters. For these experiments, we tested some of them but we also fixed a lot of others. Tuning hyperparameters is very expensive in time and computation. We can only encourage people to share their findings, results and good practices. The consequence of these variations and small differences is that good performance can be obtained with various configurations. 

In general, the various types of GAN produce good approximations and are very promising algorithms for regression task in non-linear problems. 

\chapter{Conclusions}
All these experiments are very encouraging. They show that adversarial regression produces meaningful results. However, it is necessary to study further the performance of adversarial regressions. The research about adversarial regression should be followed at least in three directions. 

First, the tests about the performance of GANs should be extended to a broader cast of experiments. More systematic analysis of high dimensional experiments should test the performances of different GANs for these type of cases. In the same idea, we can imagine experiments where we we have more than one unknown variable. 

Second, the GANs themselves should be improved in the perspective of doing regression task. So far, the research on GAN is mainly concentrated on image generation. This task does not require the same skills. To generate images a broad approximation of the main modes of the distribution is sufficient, while for regression we want to be as close as possible to the real distribution. GANs are new. They are far of being completely understood. It is a chance in the sense that we can still expect important improvement. It is possible to find new solutions to improve GANs for adversarial regression. 

Third, GANs are not the only recently developed technique able to produce high dimensional non-linear regression. The machine learning community has an intense activity and creativity. Among the most popular, Variational Auto-Encoder and Bayesian Neural Networks can also be used for high dimension non-linear regressions.

Even if adversion regression has to be studied further, we have shown promising results. We have seen that adversarial regression can produce accurate estimates for non-linear regression. It is not only able to give a point estimate but an approximation of the full probability density function for a new observation. We also have seen promising results in high dimension. Moreover, adversarial regression is in the very early stage of development. We can expect major improvment in the coming month and years. This promising results make of adversarial regression one of the most promising method for non-linear regression in high dimensional space.

\bibliographystyle{nat}

\end{document}